\documentclass[journal]{IEEEtran}
\usepackage{ifpdf}
\pdfoutput=1
\usepackage{times}
\usepackage{epsfig}
\usepackage{graphicx}
\usepackage{amsmath}
\usepackage{amssymb}
\usepackage{subfigure}
\usepackage{scrextend}
\usepackage[nocompress]{cite}
\usepackage{algorithmic}
\usepackage{color}
\usepackage{placeins}
\usepackage{url}
\usepackage{cite}

\usepackage[pagebackref=true,breaklinks=true,letterpaper=true,colorlinks,bookmarks=false]{hyperref}

\newcommand{\etal}{\emph{et al.}}
\newcommand{\eg}{\emph{e.g.,}}
\newcommand{\ie}{\emph{i.e.,}}

\begin{document}
\title{Deep Convolution Networks for Compression Artifacts Reduction}
%
%
%

\author{Ke~Yu,
        Chao~Dong,
        Chen~Change~Loy,~\IEEEmembership{Member,~IEEE,}
        Xiaoou~Tang,~\IEEEmembership{Fellow,~IEEE,}
\thanks{Ke Yu is with the Department of Electronic Engineering, Tsinghua University, Beijing, P.R.China,
 e-mail: (yk11@mails.tsinghua.edu.cn).}
\thanks{Chao Dong, Chen Change Loy (corresponding author), and Xiaoou Tang are with the Department
of Information Engineering, The Chinese University of Hong Kong,
 e-mail: ( \{dc012,ccloy,xtang\}@ie.cuhk.edu.com).}
}


\maketitle

\begin{abstract}
Lossy compression introduces complex compression artifacts, particularly blocking artifacts, ringing effects and blurring. Existing algorithms either focus on removing blocking artifacts and produce blurred output, or restore sharpened images that are accompanied with ringing effects.
Inspired by the success of deep convolutional networks (DCN) on super-resolution~\cite{Dong2014}, we formulate a compact and efficient network for seamless attenuation of different compression artifacts.
To meet the speed requirement of real-world applications, we further accelerate the proposed baseline model by layer decomposition and joint use of large-stride convolutional and deconvolutional layers. This also leads to a more general CNN framework that has a close relationship with the conventional Multi-Layer Perceptron (MLP). Finally, the modified network achieves a speed up of 7.5$\times$ with almost no performance loss compared to the baseline model.
We also demonstrate that a deeper model can be effectively trained with features learned in a shallow network. Following a similar ``easy to hard'' idea, we systematically investigate three practical transfer settings and show the effectiveness of transfer learning in low-level vision problems.
Our method shows superior performance than the state-of-the-art methods both on benchmark datasets and a real-world use case. 
%
\end{abstract}

\begin{IEEEkeywords}
Convolutional Network, Deconvolution, Compression artifacts, JPEG compression
\end{IEEEkeywords}

\IEEEpeerreviewmaketitle

\section{Introduction}
\label{sec:introduction}
\begin{figure}[t]\footnotesize
\centering
\subfigure[Left: the JPEG-compressed image, where we could see blocking artifacts, ringing effects and blurring on the eyes, abrupt intensity changes on the face. Right: the restored image by the proposed deep model (AR-CNN), where we remove these compression artifacts and produce sharp details.]{
  \label{fig:introductiona}
  \includegraphics[width=\linewidth]{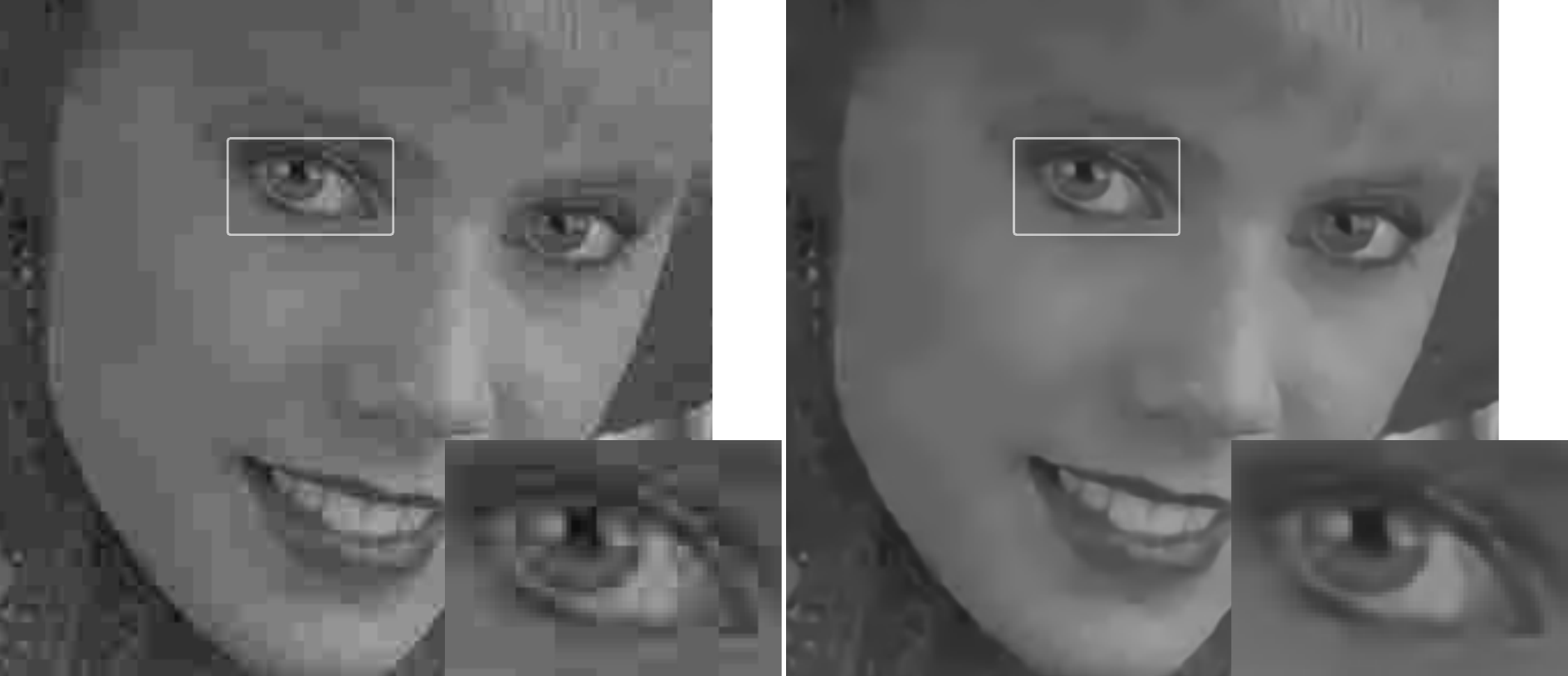}
}
\vskip -0.1cm

\subfigure[Left: the Twitter-compressed image, which is first re-scaled to a small image and then compressed on the server-side. Right: the restored image by the proposed deep model (AR-CNN)]{
  \label{fig:introductionb}
  \includegraphics[width=\linewidth]{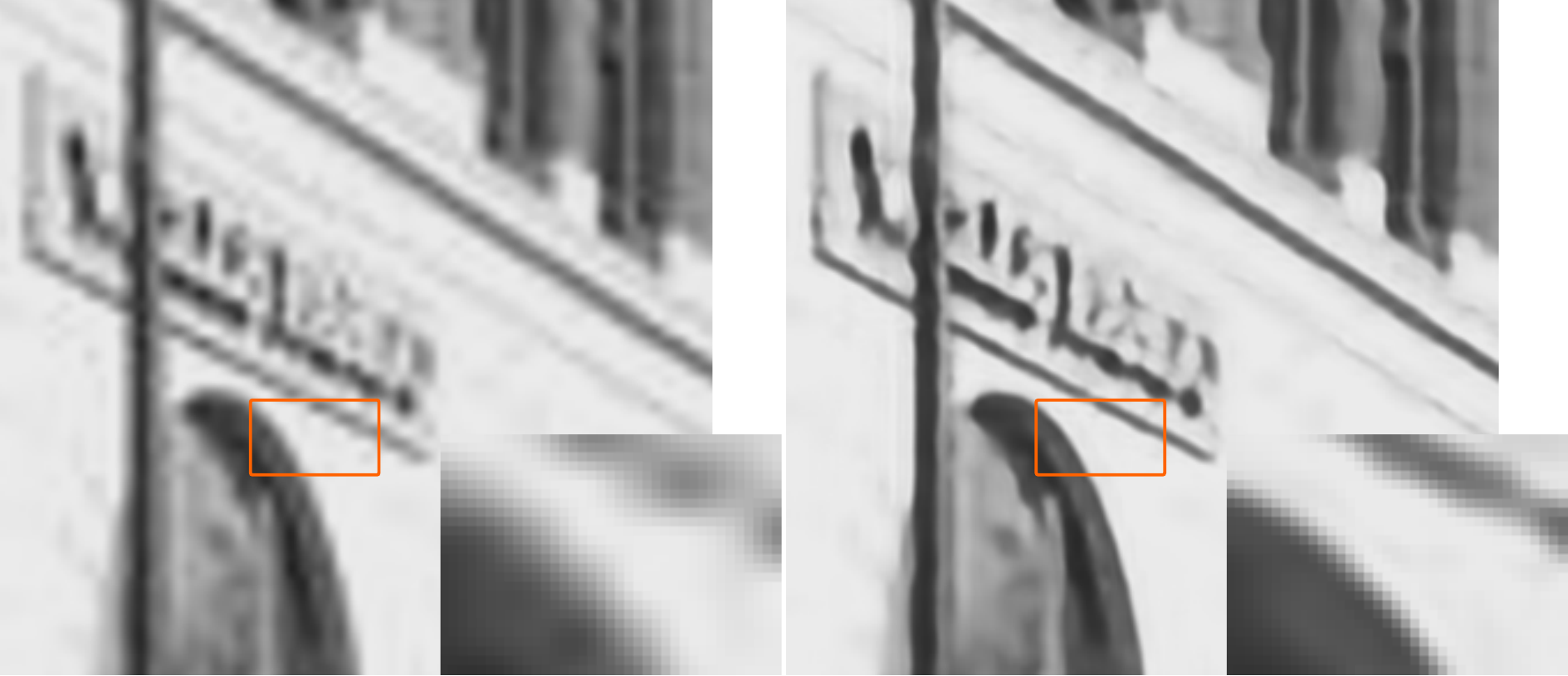}
}
\vskip -0.1cm
  \caption{Example compressed images and our restoration results on the JPEG compression scheme and the real use case -- \textit{Twitter}.}
  \label{fig:introduction}
\vskip -0.5cm
\end{figure}

Lossy compression (\eg~JPEG, WebP and HEVC-MSP) is one class of data encoding methods that uses inexact approximations for representing the encoded content. In this age of information explosion, lossy compression is indispensable and inevitable for companies (\eg~\textit{Twitter} and~\textit{Facebook}) to save bandwidth and storage space. However, compression in its nature will introduce undesired complex artifacts, which will severely reduce the user experience (\eg~Figure~\ref{fig:introduction}). All these artifacts not only decrease perceptual visual quality, but also adversely affect various low-level image processing routines that take compressed images as input, \eg~contrast enhancement~\cite{Li2014}, super-resolution~\cite{Yang2014,Dong2014}, and edge detection~\cite{Dollar2013}. Despite the huge demand, effective compression artifacts reduction remains an open problem.

Various compression schemes bring different kinds of compression artifacts, which are all complex and signal-dependent.
Take JPEG compression as an example, the discontinuities between adjacent 8$\times$8 pixel blocks will result in \textit{blocking artifacts}, while the coarse quantization of the high-frequency components will bring \textit{ringing effects} and \textit{blurring}, as depicted in Figure~\ref{fig:introductiona}.
%
As an improved version of JPEG, JPEG 2000 adopts wavelet transform to avoid blocking artifacts, but still exhibits ringing effects and blurring. Apart from the widely-adopted compression standards, commercials also introduced their own compression schemes to meet specific requirements. For example, \textit{Twitter} and~\textit{Facebook} will compress the uploaded high-resolution images by first re-scaling and then compression. The combined compression strategies also introduce severe ringing effects and blurring, but in a different manner (see Figure~\ref{fig:introductionb}).

To cope with various compression artifacts, different approaches have been proposed, some of which are designed for a specific compression standard, especially JPEG.
%
For instance, deblocking oriented approaches~\cite{List2003,ReeveIII1984,Wang2013} perform filtering along the block boundaries to reduce only blocking artifacts. Liew~\etal~\cite{Liew2004} and Foi~\etal~\cite{Foi2007} use thresholding by wavelet transform and Shape-Adaptive DCT transform, respectively.
With the help of problem-specific priors (\eg~the quantization table), Liu~\etal~\cite{Liu2015} exploit residual redundancies in the DCT domain and propose a sparsity-based dual-domain (DCT and pixel domains) approach. Wang~\etal~\cite{Wang2016} further introduce deep sparse-coding networks to the DCT and pixel domains and achieve superior performance. This kind of methods can be referred to as soft decoding for a specific compression standard (\eg~JPEG), and can be hardly extended to other compression schemes.
Alternatively, data-driven learning-based methods have better generalization ability. Jung~\etal~\cite{Jung2012} propose restoration method based on sparse representation. Kwon~\etal~\cite{Kwon2015} adopt the Gaussian process (GP) regression to achieve both super-resolution and compression artifact removal. The adjusted anchored neighborhood regression (A+) approach~\cite{Rothe2015} is also used to enhance JPEG 2000 images. These methods can be easily generalized for different tasks.

Deep learning has shown impressive results on both high-level and low-level vision problems. In particular, the Super-Resolution Convolutional Neural Network (SRCNN) proposed by Dong~\etal~\cite{Dong2014} shows the great potential of an end-to-end DCN in image super-resolution. The study also points out that conventional sparse-coding-based image restoration model can be equally seen as a deep model.
However, if we directly apply SRCNN in compression artifact reduction, the features extracted by its first layer could be noisy, leading to undesirable noisy patterns in reconstruction. Thus the three-layer SRCNN is not well suited for restoring compressed images, especially in dealing with complex artifacts.

To eliminate the undesired artifacts, we improve SRCNN by embedding one or more ``feature enhancement'' layers after the first layer to clean the noisy features. Experiments show that the improved model, namely Artifacts Reduction Convolutional Neural Networks (AR-CNN), is exceptionally effective in suppressing blocking artifacts while retaining edge patterns and sharp details (see Figure~\ref{fig:introduction}). Different from the JPEG-specific models, AR-CNN is equally effective in coping with different compression schemes, including JPEG, JPEG 2000, \textit{Twitter} and so on.

However, the network scale increases significantly when we add another layer, making it hard to be applied in real-world applications. Generally, the high computational cost has been a major bottleneck for most previous methods~\cite{Wang2016}. When delving into the network structure, we find two key factors that restrict the inference speed. First, the added ``feature enhancement'' layer accounts for almost $95\%$ of the total parameters. Second, when we adopt a fully-convolution structure, the time complexity will increase quadratically with the spatial size of the input image.

To accelerate the inference process while still maintaining good performance, we investigate a more efficient framework with two main modifications. For the redundant parameters, we insert another ``shrinking'' layer with $1\times 1$ filters between the first two layers. For the large computation load of convolution, we use large-stride convolution filters in the first layer and the corresponding deconvolution filters in the last layer. Then the convolution operation in the middle layers will be conducted on smaller feature maps, leading to much faster inference. Experiments show that the modified network, namely Fast AR-CNN, can be 7.5 times faster than the baseline AR-CNN with almost no performance loss. This further helps us formulate a more general CNN framework for low-level vision problems. We also reveal its close relationship with the conventional Multi-Layer Perceptron~\cite{Burger2012}.

Another issue we met is how to effectively train a deeper DCN.
As pointed out in SRCNN~\cite{Dong2014a}, training a five-layer network becomes a bottleneck. The difficulty of training is partially due to the sub-optimal initialization settings.
The aforementioned difficulty motivates us to investigate a better way to train a deeper model for low-level vision problems. We find that this can be effectively solved by transferring the features learned in a shallow network to a deeper one and fine-tuning simultaneously\footnote{Generally, the transfer learning method will train a base network first, and copy the learned parameters or features of several layers to the corresponding layers of a target network. These transferred layers can be left frozen or fine-tuned to the target dataset. The remaining layers are randomly initialized and trained to the target task.}. This strategy has also been proven successful in learning a deeper CNN for image classification~\cite{Simonyan2014}.
Following a similar general intuitive idea, \textit{easy to hard}, we discover other interesting transfer settings in our low-level vision task:
(1) We transfer the features learned in a high-quality compression model (easier) to a low-quality one (harder), and find that it converges faster than random initialization.
(2) In the real use case, companies tend to apply different compression strategies (including re-scaling) according to their purposes (\eg~Figure~\ref{fig:introductionb}). We transfer the features learned in a standard compression model (easier) to a real use case (harder), and find that it performs better than learning from scratch.

The contributions of this study are four-fold:
(1) We formulate a new deep convolutional network for efficient reduction of various compression artifacts.
Extensive experiments, including that on real use cases,
demonstrate the effectiveness of our method over state-of-the-art methods~\cite{Foi2007} both perceptually and quantitatively.
(2) We progressively modify the baseline model AR-CNN and present a more efficient network structure, which achieves a speed up of $7.5\times$ compared to the baseline AR-CNN while still maintaining the state-of-the-art performance.
(3) We verify that reusing the features in shallow networks is helpful in learning a deeper model for compression artifacts reduction. Under the same intuitive idea -- \textit{easy to hard}, we reveal a number of interesting and practical transfer settings.
%

The preliminary version of this work was published earlier~\cite{Dong2015}. In this work, we make significant improvements in both methodology and experiments. First, in the methodology, we add analysis on the computational cost of the proposed model, and point out two key factors that affect the time efficiency. Then we propose the corresponding acceleration strategies, and extend the baseline model to a more general and efficient network structure. In the experiments, we adopt data augmentation to further push the performance. In addition, we conduct experiments on JPEG 2000 images and show superior performance to the state-of-the-art methods~\cite{Rothe2015,Kwon2015,Roth2009}. A detailed investigation of network settings of the new framework is presented afterwards. 

\begin{figure*}[t]\small
\begin{center}
 \includegraphics[width=\linewidth]{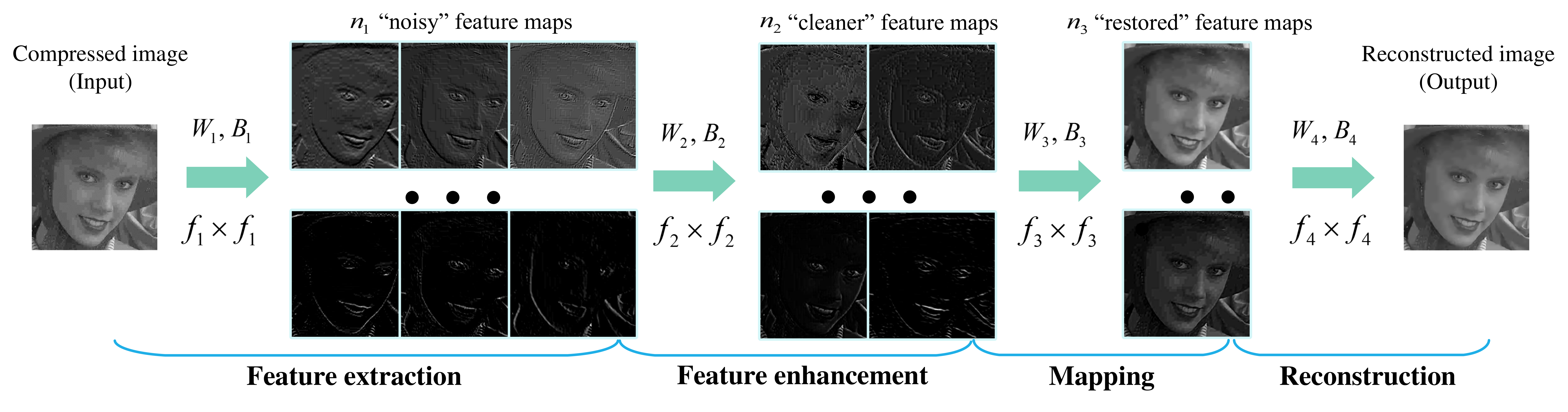}
\caption{The framework of the Artifacts Reduction Convolutional Neural Network (AR-CNN). The network consists of four convolutional layers, each of which is responsible for a specific operation. Then it optimizes the four operations (\ie~feature extraction, feature enhancement, mapping and reconstruction) jointly in an end-to-end framework. Example feature maps shown in each step could well illustrate the functionality of each operation. They are normalized for better visualization.}
\label{fig:framework}
\vspace{-0.5cm}
\end{center}
\end{figure*}

\section{Related work}

Existing algorithms can be classified into deblocking oriented and restoration oriented methods.
The deblocking oriented methods focus on removing blocking and ringing artifacts.
In the spatial domain, different kinds of filters~\cite{List2003,ReeveIII1984,Wang2013} have been proposed to adaptively deal with blocking artifacts in specific regions (\eg~edge, texture, and smooth regions). In the frequency domain, Liew~\etal~\cite{Liew2004} utilize wavelet transform and derive thresholds at different wavelet scales for denoising. The most successful deblocking oriented method is perhaps the Pointwise Shape-Adaptive DCT (SA-DCT)~\cite{Foi2007}, which is widely acknowledged as the state-of-the-art approach~\cite{Jancsary2012,Li2014}.
However, as most deblocking oriented methods, SA-DCT could not reproduce sharp edges, and tend to overly smooth texture regions.

The restoration oriented methods regard the compression operation as distortion and aim to reduce such distortion. These methods include projection on convex sets based method (POCS)~\cite{Yang1995}, solving an MAP problem (FoE)~\cite{Sun2007}, sparse-coding-based method~\cite{Jung2012}, semi-local Gassian process model~\cite{Kwon2015}, the Regression Tree Fields based method (RTF)~\cite{Jancsary2012} and adjusted anchored neighborhood regression (A+)~\cite{Rothe2015}. The RTF takes the results of SA-DCT~\cite{Foi2007} as bases and produces globally consistent image reconstructions with a regression tree field model.
It could also be optimized for any differentiable loss functions (\eg~SSIM), but often at the cost of performing sub-optimally on other evaluation metrics. As a recent method for image super-resolution~\cite{Timofte2014}, A+~\cite{Rothe2015} has also been successfully applied for compression artifacts reduction. In their method, the input image is decomposed into overlapping patches and sparsely represented by a dictionary of anchoring points. Then the uncompressed patches are predicted by multiplying with the corresponding linear regressors. They obtain impressive results on JPEG 2000 image, but have not tested on other compression schemes.

To deal with a specific compression standard, specially JPEG, some recent progresses incorporate information from dual-domains (DCT and pixel domains) and achieve impressive results. Specifically, Liu~\etal~\cite{Liu2015} apply sparse-coding in the DCT-domain to eliminate the quantization error, then restore the lost high frequency components in the pixel domain. On their basis, Wang~\etal~\cite{Wang2016} replace the sparse-coding steps with deep neural networks in both domains and achieve superior performance. These methods all require the problem-specific prior knowledge (\eg~the quantization table) and process on the $8\times8$ pixel blocks, thus cannot be generalized to other compression schemes, such as JPEG 2000 and \textit{Tiwtter}.

Super-Resolution Convolutional Neural Network (SRCNN)~\cite{Dong2014} is closely related to our work. In the study, independent steps in the sparse-coding-based method are formulated as different convolutional layers and optimized in a unified network. It shows the potential of deep model in low-level vision problems like super-resolution. However, the problem of compression is different from super-resolution in that the former consists of different kinds of artifacts. Designing a deep model for compression restoration requires a deep understanding into the different artifacts. We show that directly applying the SRCNN architecture for compression restoration will result in undesired noisy patterns in the reconstructed image.

Transfer learning in deep neural networks becomes popular since the success of deep learning in image classification~\cite{Krizhevsky2012}. The features learned from the ImageNet show good generalization ability~\cite{Zeiler2014} and become a powerful tool for several high-level vision problems, such as Pascal VOC image classification~\cite{Oquab2014} and object detection~\cite{Girshick2014,Sermanet2013}.
Yosinski~\etal~\cite{Yosinski2014} have also tried to quantify the degree to which a particular layer is general or specific. Overall, transfer learning has been systematically investigated in high-level vision problems, but not in low-level vision tasks. In this study, we explore several transfer settings on compression artifacts reduction and show the effectiveness of transfer learning in low-level vision problems.

\section{Methodology}
\label{sec:Methodology}

Our proposed approach is based on the current successful low-level vision model -- SRCNN~\cite{Dong2014}. To have a better understanding of our work, we first give a brief overview of SRCNN. Then we explain the insights that lead to a deeper network and present our new model. Subsequently, we explore three types of transfer learning strategies that help in training a deeper and better network.

\subsection{Review of SRCNN}
\label{sec:SRCNN}
The SRCNN aims at learning an end-to-end mapping, which takes the low-resolution image $\mathbf{Y}$ (after interpolation) as input and directly outputs the high-resolution one $F(\mathbf{Y})$. The network contains three convolutional layers, each of which is responsible for a specific task. Specifically, the first layer performs \textbf{patch extraction and representation}, which extracts overlapping patches from the input image and represents each patch as a high-dimensional vector. Then the \textbf{non-linear mapping} layer maps each high-dimensional vector of the first layer to another high-dimensional vector, which is conceptually the representation of a high-resolution patch. At last, the \textbf{reconstruction} layer aggregates the patch-wise representations to generate the final output.
The network can be expressed as:
\begin{align}
\label{eqn:SRCNN}
F_{0}(\mathbf{Y})&=\mathbf{Y};\\
F_{i}(\mathbf{Y})&=\max\left(0, W_{i}*F_{i-1}(\mathbf{Y})+B_{i}\right), i\in\{1,2\}; \\ F(\mathbf{Y})&=W_3*F_{2}(\mathbf{Y})+B_3,
\end{align}
%
where $W_{i}$ and $B_{i}$ represent the filters and biases of the $i$th layer respectively, $F_{i}$ is the output feature maps and ``$*$'' denotes the convolution operation. The $W_{i}$ contains $n_i$ filters of support $n_{i-1}\times f_i \times f_i$, where $f_i$ is the spatial support of a filter, $n_i$ is the number of filters, and $n_0$ is the number of channels in the input image. Note that there is no pooling or full-connected layers in SRCNN, so the final output $F(\mathbf{Y})$ is of the same size as the input image.
Rectified Linear Unit (ReLU, $\max(0,x)$)~\cite{Nair2010} is applied on the filter responses.

These three steps are analogous to the basic operations in the sparse-coding-based super-resolution methods~\cite{Yang2010a}, and this close relationship lays theoretical foundation for its successful application in super-resolution. Details can be found in the paper~\cite{Dong2014}.

\subsection{Convolutional Neural Network for Compression Artifacts Reduction}
\label{sec:ARCNN}
\textbf{Insights.} In sparse-coding-based methods and SRCNN, the first step -- feature extraction -- determines what should be emphasized and restored in the following stages. However, as various compression artifacts are coupled together, the extracted features are usually noisy and ambiguous for accurate mapping. In the experiments of reducing JPEG compression artifacts (see Section~\ref{sec:exp_SRCNN}), we find that some quantization noises coupled with high frequency details are further enhanced, bringing unexpected noisy patterns around sharp edges. Moreover, blocking artifacts in flat areas are misrecognized as normal edges, causing abrupt intensity changes in smooth regions. Inspired by the feature enhancement step in super-resolution~\cite{Xiong2009}, we introduce a feature enhancement layer after the feature extraction layer in SRCNN to form a new and deeper network -- AR-CNN. This layer maps the ``noisy'' features to a relatively ``cleaner'' feature space, which is equivalent to denoising the feature maps.

\textbf{Formulation.} The overview of the new network AR-CNN is shown in Figure~\ref{fig:framework}. The three layers of SRCNN remain unchanged in the new model. To conduct feature enhancement,
we extract new features from the $n_1$ feature maps of the first layer, and combine them to form another set of feature maps.
Overall, the AR-CNN consists of four layers, namely the feature extraction, feature enhancement, mapping and reconstruction layer.

Different from SRCNN that adopts ReLU as the activation function, we use Parametric Rectified Linear Unit (PReLU)~\cite{He2015} in the new networks. To distinguish ReLU and PReLU, we define a general activation function as:
\begin{equation}
\label{eqn:PReLU}
PReLU(x_j)=max(x_j,0)+a_j\cdot min(0,x_j),
\end{equation}
where $x_j$ is the input signal of the activation $f$ on the $j$-th channel, and $a_j$ is the coefficient of the negative part. The parameter $a_j$ is set to be zero for ReLU, but is learnable for PReLU. We choose PReLU mainly to avoid the ``dead features''~\cite{Zeiler2014} caused by zero gradients in ReLU. We represent the whole network as:
\begin{align}
\label{eqn:ARCNN}
F_{0}(\mathbf{Y})&=\mathbf{Y};\\
F_{i}(\mathbf{Y})&=PReLU \left(W_{i}*F_{i-1}(\mathbf{Y})+B_{i}\right), i\in\{1,2,3\}; \\
F(\mathbf{Y})&=W_4*F_{3}(\mathbf{Y})+B_4.
\end{align}
where the meaning of the variables is the same as that in Equation~\ref{eqn:SRCNN}, and the second layer ($W_2,B_2$) is the added feature enhancement layer.

It is worth noticing that AR-CNN is not equal to a deeper SRCNN that contains more than one non-linear mapping layers\footnote{Adding non-linear mapping layers has been suggested as an extension of SRCNN in~\cite{Dong2014}.}. A deeper SRCNN imposes more non-linearity in the mapping stage, which equals to adopting a more robust regressor between the low-level features and the final output. Similar ideas have been proposed in some sparse-coding-based methods~\cite{Kim2010,Bevilacqua2012}. However, as compression artifacts are complex, low-level features extracted by a single layer are noisy. Thus the performance bottleneck lies on the features but not the regressor. AR-CNN improves the mapping accuracy by enhancing the extracted low-level features, and the first two layers together can be regarded as a better feature extractor. This leads to better performance than a deeper SRCNN. Experimental results of AR-CNN, SRCNN and deeper SRCNN will be shown in Section~\ref{sec:exp_SRCNN}.

\subsection{Model Learning}
\label{subsec:learning}
Given a set of ground truth images $\left\{ \mathbf{X}_i\right\}$ and their corresponding compressed images $\left\{ \mathbf{Y}_i\right\}$, we use Mean Squared Error (MSE) as the loss function:
\begin{equation}
\label{eqn:loss}
L(\Theta)=\frac{1}{n}\sum_{i=1}^n||F(\mathbf{Y}_i ; \Theta) - \mathbf{X}_i||^2,
\end{equation}
where $\Theta=\{W_1,W_2,W_3,W_4,B_1,B_2,B_3,B_4\}$, $n$ is the number of training samples. The loss is minimized using stochastic gradient descent with the standard backpropagation.
We adopt a batch-mode learning method with a batch size of 128.

\section{Accelerating AR-CNN}
\label{subsec:Acceleration}
Although AR-CNN is already much smaller than most of the existing deep models (\eg~AlexNet~\cite{Krizhevsky2012} and Deepid-net~\cite{Ouyang2015}), it is still unsatisfactory for practical or even real-time on-line applications. Specifically, with an additional layer, AR-CNN has been several times larger than SRCNN in the network scale.
In this section, we progressively accelerate the proposed baseline model while preserving its reconstruction quality.
First, we analyze the computational complexity of AR-CNN and find out the most influential factors. Then we re-design the network by layer decomposition and joint use of large-stride convolutional and deconvolutional layers. We further make it a more general framework, and compare it with the conventional Multi-Layer Perceptron (MLP).

\begin{table}[b]\footnotesize
\caption{Analysis of network parameters in AR-CNN.}\label{tab:para}
\begin{center}
\begin{tabular}{|c|c|c|c|c|c|}
\hline
layer No. &  1 &  2 & 3 & 4 & total \\
\hline
Number & 5184 & 100,352 & 512 & 400 & 106,448 \\
\hline
Percentage & 4.87$\%$ &94.27$\%$ & 0.48$\%$ & 0.38$\%$ & 100$\%$ \\
\hline
\end{tabular}
\end{center}
\end{table}

\subsection{Complexity Analysis}

As AR-CNN consists of purely convolutional layers, The total number of parameters can be calculated as:

\begin{equation}
\label{eqn:para}
N=\sum_{i=1}^d n_{i-1}\cdot n_{i}\cdot f_i^2.
\end{equation}
where $i$ is the layer index, $d$ is the number of layers and $f_{i}$ is the spatial size of the filters. The number of filters of the $i$-th layer is denoted by $n_{i}$, and the number of input channels is $n_{i-1}$.
If we include the spatial size of the output feature maps $m_i$, we obtain the expression for time complexity:
\begin{equation}
\label{eqn:time}
O\{ \sum_{i=1}^d n_{i-1}\cdot n_{i}\cdot f_i^2 \cdot m_i^2\},
\end{equation}

For our baseline model AR-CNN, we set $d=4$, $n_0=1$, $n_1=64$, $n_2=32$, $n_3=16$, $n_4=1$, $f_1=9$, $f_2=7$, $f_3=1$, $f_4=5$, namely 64(9)-32(7)-16(1)-1(5). First, we analyze the parameters of each layer in Table~\ref{tab:para}.
We find that the ``feature enhancement'' layer accounts for almost $95\%$ of total parameters. Obviously, if we want to reduce the parameters, the second layer should be the breakthrough point.

On the other hand, the spatial size of the output feature maps $m_i$ also plays an important role in the overall time complexity (see Equation~\ref{eqn:time}). In conventional low-level vision models like SRCNN, the spatial size of all intermediate feature maps remains the same as that of the input image. However, this is not the case for high-level vision models like AlexNet~\cite{Krizhevsky2012}, which consists of some large-stride (stride $>1$) convolution filters. Generally, a reasonable larger stride can significantly speed up the convolution operation with little cost on accuracy, thus the stride size should be another key factor to improve our network. Based on the above observations, we explore a more efficient network structure in the next subsection.


\subsection{Acceleration Strategies}
\label{sec:accstrategy}

\begin{figure}[t]\small
\centering
\subfigure[When the stride is 1, the convolution and deconvolution can be regarded as equivalent operations. Each output pixel is determined by the same number of input pixels (in the orange circle) in convolution and deconvolution.]{
  \label{fig:deconv1}
  \includegraphics[width=\linewidth]{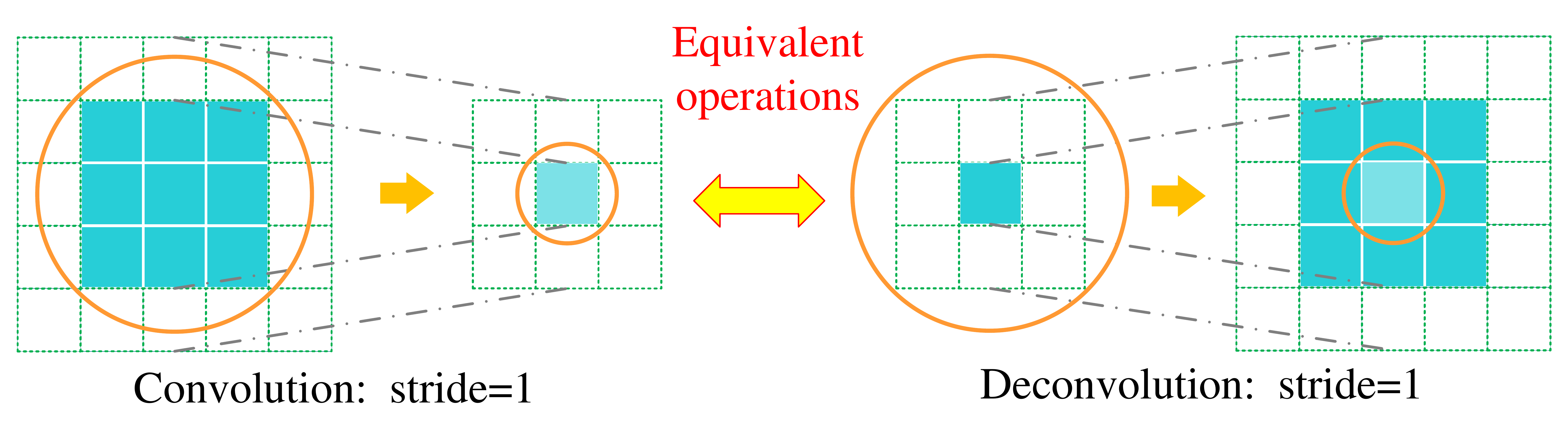}
}
\subfigure[When the stride is larger than 1, the convolution performs downsampling, and the deconvolution performs upsampling.]{
  \label{fig:deconv2}
  \includegraphics[width=\linewidth]{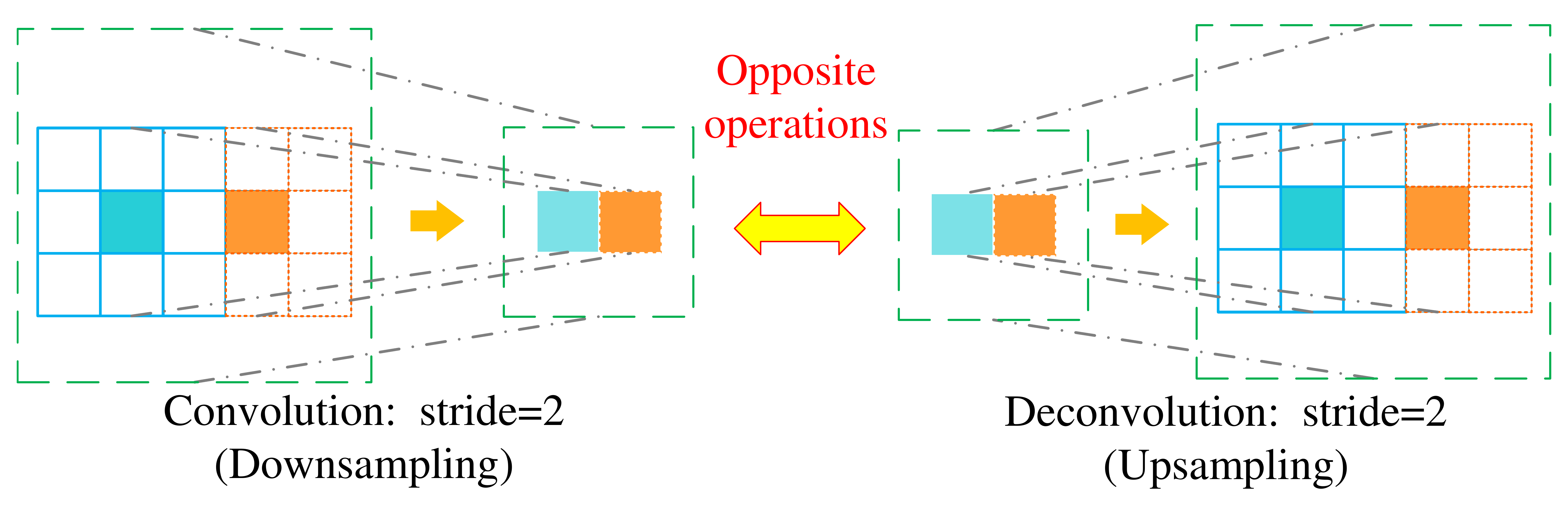}
}
\caption{The illustration of convolution and deconvolution process.}
\label{fig:deconv}
\vskip -0.5cm
\end{figure}
\begin{figure*}[t]\small
\begin{center}

 \includegraphics[width=0.75\linewidth]{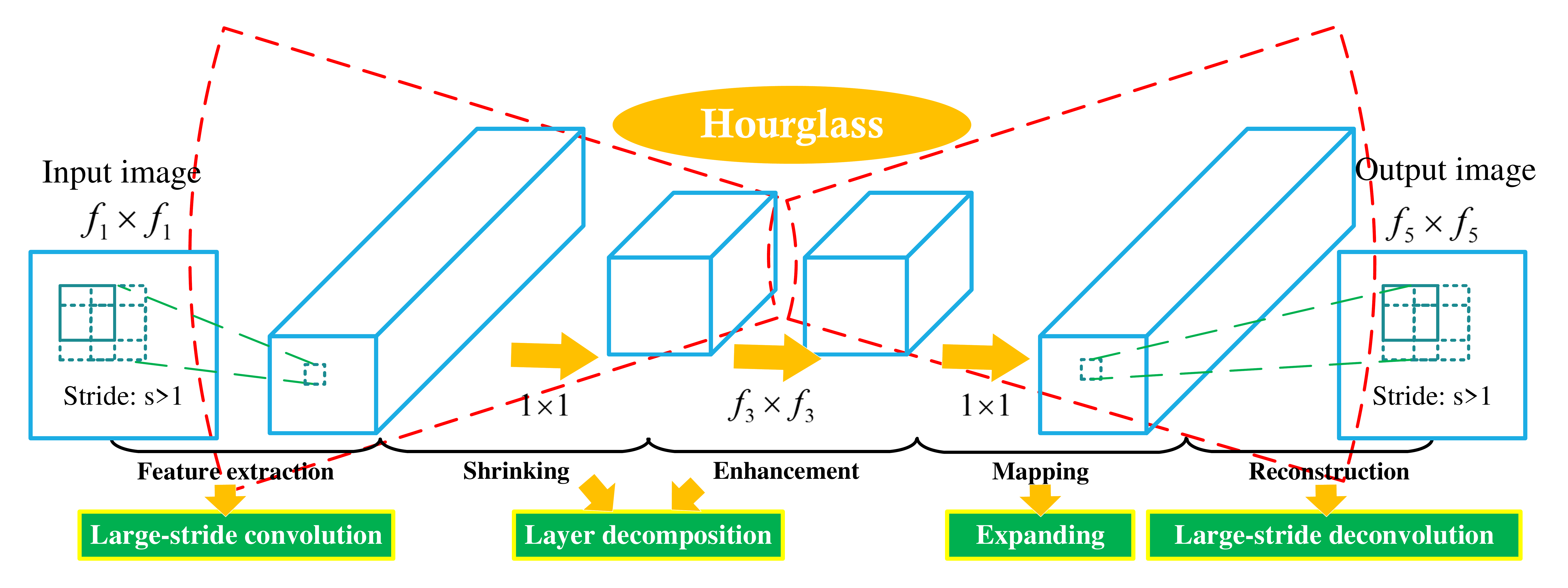}
\caption{The framework of the Fast AR-CNN. There are two main modifications based on the original AR-CNN. First, the layer decomposition splits the original ``feature enhancement'' layer into a ``shrinking'' layer and an ``enhancement'' layer. Then the large-stride convolutional and deconvolutional layers significantly decrease the spatial size of the feature maps of the middle layers. The overall shape of the framework is like an hourglass, which is thick at the ends and thin in the middle.}
\label{fig:framework2}
\vspace{-0.5cm}
\end{center}
\end{figure*}

\textbf{Layer decomposition.} We first reduce the complexity of the ``feature enhancement'' layer. This layer plays two roles simultaneously. One is to denoise the input feature maps with a set of large filters (\ie~$7\times 7$), and the other is to map the high dimensional features to a relatively low dimensional feature space (\ie~from 64 to 32). This indicates that we can replace it with two connected layers, each of which is responsible for a single task. To be specific, we decompose the ``feature enhancement'' layer into a ``shrinking'' layer with 32 $1\times 1$ filters and an ``enhancement'' layer with 32 $7\times 7$ filters, as shown in Figure~\ref{fig:framework2}. Note that the $1\times 1$ filters are widely used to reduce the feature dimensions in deep models~\cite{Lin2014}. Then we can calculate the parameters as follows:

\begin{equation}
\label{eqn:time}
32\cdot 7^2\cdot 64 = 100,352 \\
\rightarrow 32\cdot 1^2\cdot 32 + 32\cdot 7^2\cdot 32 = 51,200.
\end{equation}

It is clear that the parameters are reduced almost by half. Correspondingly, the overall network scale also decreases by $46.17\%$. We denote the modified network as 64(9)-32(1)-32(7)-16(1)-1(5). In Section \ref{sec:settings1}, we will show that this model achieves almost the same restoration quality as the baseline model 64(9)-32(7)-16(1)-1(5).

\textbf{Large-stride convolution and deconvolution.} Another acceleration strategy is to increase the stride size (\eg~stride $s>1$) in the first convolutional layer.
In AR-CNN, the first layer plays a similar role (\ie~feature extractor) as in high-level vision deep models, thus it is a worthy attempt to increase the stride size, \eg~from 1 to 2.

However, this will result in a smaller output and affect the end-to-end mapping structure. To address this problem, we replace the last convolutional layer of AR-CNN (Figure~\ref{fig:framework}) with a deconvolutional layer. The deconvolution can be regarded as an opposite operation of convolution. Specially, if we set the stride $s=1$, the function of a deconvolution filter is equal to that of a convolution filter (see Figure~\ref{fig:deconv1}). For a larger stride $s>1$, the convolution performs sub-sampling, while the deconvolution performs up-sampling (see Figure~\ref{fig:deconv2}). Therefore, if we use the same stride for the first and the last layer, the output will remain the same size as the input, as depicted in Figure~\ref{fig:framework2}. After joint use of large-stride convolutional and deconvolutional layers, the spatial size of the feature maps $m_i$ will become $m_i/s$, which will reduce the overall time complexity significantly.

Although the above modifications will improve the time efficiency, they may also influence the restoration quality.
To further improve the performance, we can expand the mapping layer (\ie~use more mapping filters) and enlarge the filter size of the deconvolutional layer. For instance, we can set the number of mapping filters to be same as that of the first-layer filters (\ie~from 16 to 64), and use the same filter size for the first and the last layer (\ie~$f_1=f_5=9$). This is a feasible solution but not a strict rule. In general, it can be seen as a compensation for the low time complexity. In Section~\ref{sec:settings1}, we investigate different settings through a series of controlled experiments, and find a good trade-off between performance and complexity.



\textbf{Fast AR-CNN.} Through the above modifications, we reach to a more efficient network structure. If we set $s=2$, the modified model can be represented as 64(9)-32(1)-32(7)-64(1)-1[9]-s2, where the square bracket refers to the deconvolution filter. We name the new model as \textit{Fast AR-CNN}. The number of its overall parameters is 56,496 by Equation~\ref{eqn:para}. Then the acceleration ratio can be calculated as $106448/56496\cdot 2^2=7.5$. Note that this network could achieve similar results as the baseline model as shown in Section \ref{sec:settings1}.

\subsection{A General Framework}
\label{sec:general}
When we relax the network settings, such as the filter number, filter size, and stride, we can obtain a more general framework with some appealing properties as follows.

(1) The overall ``shape'' of the network is like an ``hourglass'', which is thick at the ends and thin in the middle. The shrinking and the mapping layers control the width of the network. They are all $1\times 1$ filters and contribute little to the overall complexity.

(2) The choice of the stride can be very flexible. The previous low-level vision CNNs, such as SRCNN and AR-CNN, can be seen as a special case of $s=1$, where the deconvolutional layer is equal to a convolutional layer. When $s>1$, the time complexity will decrease $s^2$ times at the cost of the reconstruction quality.

(3) When we adopt all $1\times 1$ filters in the middle layer, it will work very similar to a Multi-Layer Perception (MLP)~\cite{Burger2012}. The MLP processes each patch individually. Input patches are extracted from the image with a stride $s$, and the output patches are aggregated (\ie~averaging) on the overlapping areas. While for our framework, the patches are also extracted with a stride $s$, but in a convolution manner. The output patches are also aggregated (\ie~summation) on overlapping areas, but in a deconvolution manner. If the filter size of the middle layers is set to 1, then each output patch is determined purely by a single input patch, which is almost the same as a MLP. However, when we set a larger filter size for middle layers, the receptive field of an output patch will increase, leading to much better performance. This also reveals why the CNN structure can outperform the conventional MLP theoretically.

Here, we present the general framework as
\begin{equation}
\label{eqn:general}
n_1(f_1)-n_2(1)-n_3(f_3)\times m-n_4(1)-n_5[f_5]-s,
\end{equation}
where $f$ and $n$ represent the filter size and the number of filters respectively. The number of middle layers is denoted as $m$, and can be used to design a deeper network. As we focus more on speed, we just set $m=1$ in the following experiments. Figure~\ref{fig:framework2} shows the overall structure of the new framework. We believe that this framework can be applied to more low-level vision problems, such as denoising and deblurring, but this is beyond the scope of this paper.

\section{Easy-Hard Transfer}

\begin{figure}[t]\small
\begin{center}
\vskip -0.5cm
 \includegraphics[width=1\linewidth]{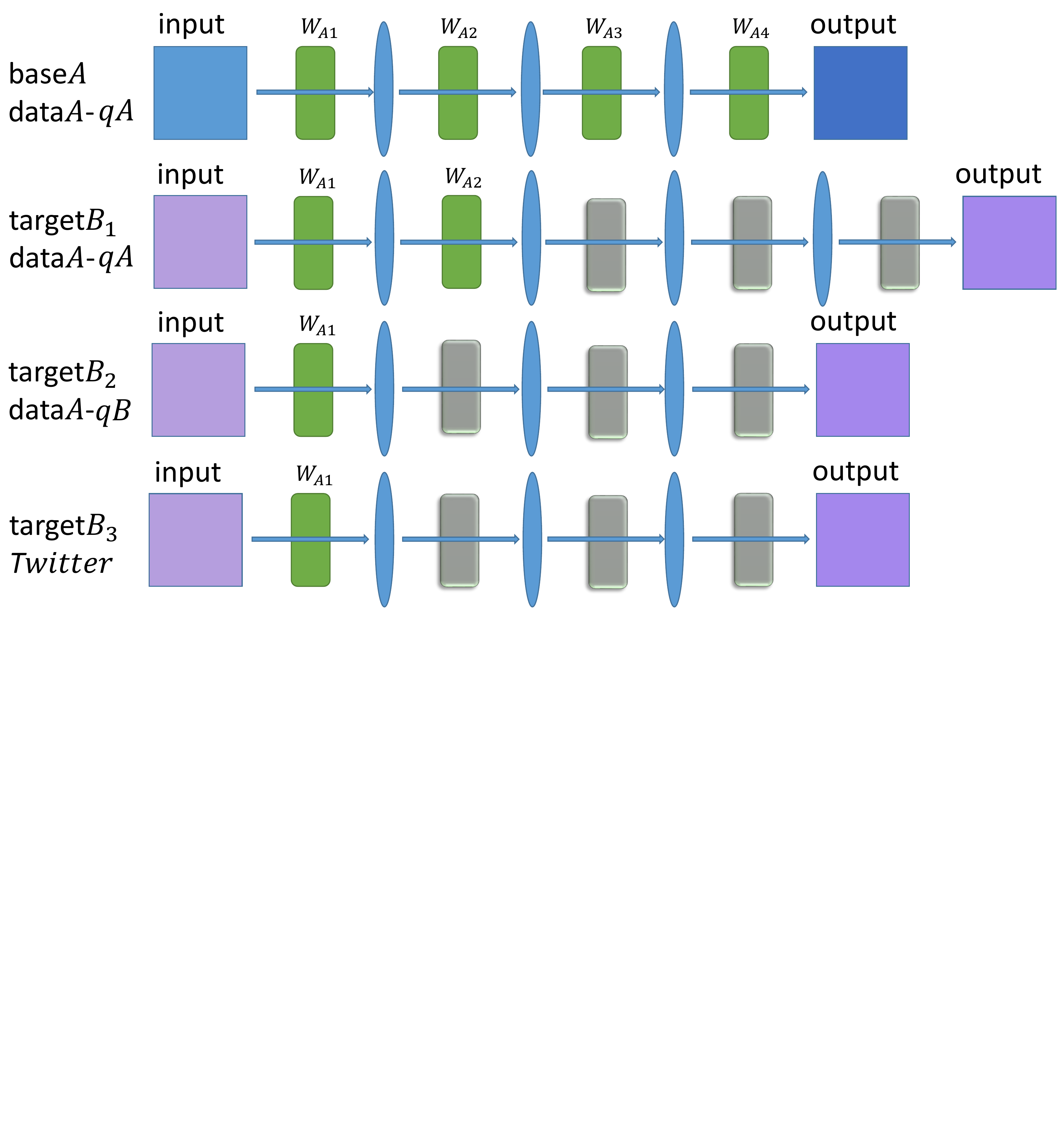}
\vskip -4.4cm
\caption{Easy-hard transfer settings. First row: The baseline 4-layer network trained with \textit{dataA}-$qA$. Second row: The 5-layer AR-CNN targeted at \textit{dataA}-$qA$. Third row: The AR-CNN targeted at \textit{dataA}-$qB$. Fourth row: The AR-CNN targeted at \textit{Twitter} data. Green boxes indicate the transferred features from the base network, and gray boxes represent random initialization. The ellipsoidal bars between weight vectors represent the activation functions.}
\label{fig:Easy-hard1}
\vspace{-0.5cm}
\end{center}
\end{figure}

Transfer learning in deep models provides an effective way of initialization. In fact, conventional initialization strategies (\ie~randomly drawn from Gaussian distributions with fixed standard deviations~\cite{Krizhevsky2012}) are found not suitable for training a very deep model, as reported in~\cite{He2015}.
To address this issue, He~\etal~\cite{He2015} derive a robust initialization method for rectifier nonlinearities,  Simonyan~\etal~\cite{Simonyan2014} propose to use the pre-trained features on a shallow network for initialization.


In low-level vision problems (\eg~super resolution), it is observed that training a network beyond 4 layers would encounter the problem of convergence, even that a large number of training images (\eg~ImageNet) are provided~\cite{Dong2014}.
We are also met with this difficulty during the training process of AR-CNN.
To this end, we systematically investigate several transfer settings in training a low-level vision network following an intuitive idea of ``easy-hard transfer''.
Specifically, we attempt to reuse the features learned in a relatively easier task to initialize a deeper or harder network.
Interestingly, the concept ``easy-hard transfer'' has already been pointed out in neuro-computation study~\cite{Gluck1993}, where the prior training on an easy discrimination can help learn a second harder one.



Formally, we define the base (or source) task as \textit{A} and the target tasks as \textit{$B_i$}, $i\in \{1,2,3\}$. As shown in Figure~\ref{fig:Easy-hard1}, the base network \textit{baseA} is a four-layer AR-CNN trained on a large dataset \textit{dataA}, of which images are compressed using a standard compression scheme with the compression quality \textit{qA}. All layers in \textit{baseA} are randomly initialized from a Gaussian distribution. We will transfer one or two layers of \textit{baseA} to different target tasks (see Figure~\ref{fig:Easy-hard1}). Such transfers can be described as follows.

\textbf{Transfer shallow to deeper model.}
As indicated by~\cite{Dong2014a}, a five-layer network is sensitive to the initialization parameters and learning rate. Thus we transfer the first two layers of \textit{baseA} to a five-layer network \textit{target$B_1$}. Then we randomly initialize its remaining layers\footnote{Random initialization on remaining layers are also applied similarly for tasks \textit{$B_2$}, and \textit{$B_3$}.} and train all layers toward the same dataset \textit{dataA}. This is conceptually similar to that applied in image classification~\cite{Simonyan2014}, but this approach has never been validated in low-level vision problems.

\textbf{Transfer high to low quality.}
Images of low compression quality contain more complex artifacts. Here we use the features learned from high compression quality images as a starting point to help learn more complicated features in the DCN. Specifically, the first layer of \textit{target$B_2$} are copied from \textit{baseA} and trained on images that are compressed with a lower compression quality \textit{qB}.


\textbf{Transfer standard to real use case.}
We then explore whether the features learned under a standard compression scheme can be generalized to other real use cases, which often contain more complex artifacts due to different levels of re-scaling and compression. We transfer the first layer of \textit{baseA} to the network \textit{target$B_3$}, and train all layers on the new dataset.

\begin{figure}[t]\small
\centering
\subfigure[High compression quality (quality 20 in MATLAB encoder)]{
  \label{pattern_q20}
  \includegraphics[width=\linewidth]{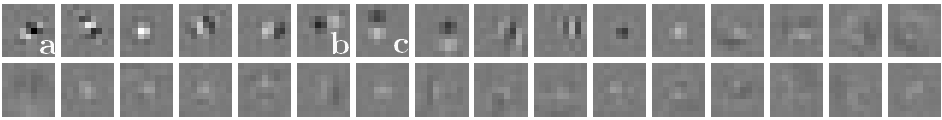}
}
\subfigure[Low compression quality (quality 10 in MATLAB encoder)]{
  \label{pattern_q10}
  \includegraphics[width=\linewidth]{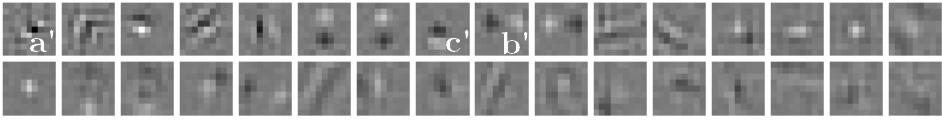}
}
  \caption{First layer filters of AR-CNN learned under different JPEG compression qualities.}
  \label{fig:features}
\vskip -0.5cm
\end{figure}

\textbf{Discussion.}
%
%
Why are the features learned from relatively easy tasks helpful? First, features from a well-trained network can provide a good starting point. Then the rest of a deeper model can be regarded as shallow one, which is easier to converge. Second, features learned in different tasks always have a lot in common. For instance, Figure~\ref{fig:features} shows the features learned under different JPEG compression qualities. Obviously, filters $a, b, c$ of high quality are very similar to filters $a', b', c'$ of low quality. This kind of features can be reused or improved during fine-tuning, making the convergence faster and more stable. Furthermore, a deep network for a hard problem can be seen as an insufficiently biased learner with overly large hypothesis space to search, and therefore is prone to overfitting. These few transfer settings we investigate introduce good bias to enable the learner to acquire a concept with greater generality. Experimental results in Section~\ref{sec:transfer} validate the above analysis.

\section{Experiments}
\label{subsec:settings}

We use the BSDS500 dataset~\cite{Arbelaez2011} as our training set. Specifically, its disjoint training set (200 images) and test set (200 images) are all used for training, and its validation set (100 images) is used for validation.
To use the dataset more efficiently, we adopt data augmentation for the training images in two steps. 1) Scaling: each image is scaled by a factor of 0.9, 0.8, 0.7 and 0.6. 2) Rotation: each image is rotated by a degree of 90, 180 and 270. Then our augmented training set is $5 \times 4 = 20$ times of the original one.
We only focus on the restoration of the luminance channel (in YCrCb space) in this paper.

The training image pairs $\left\{Y, X \right\}$ are prepared as follows. Images in the training set are decomposed into $24\times 24$ sub-images\footnote{We use sub-images because we regard each sample as an image rather than a big patch.} $X = \{\mathbf{X}_i\}_{i=1}^n$. Then the compressed samples $Y=\{\mathbf{Y}_i\}_{i=1}^n$ are generated from the training samples. The sub-images are extracted from the ground truth images with a stride of 20. Thus the augmented $400 \times 20 = 8000$ training images could provide 1,870,336 training samples. We adopt zero padding for the layers with a filter size larger than 1. As the training is implemented with the \textit{Caffe} package~\cite{Jia2014}, the deconvolution filter will output a feature map with ($s-1$)-pixel cut on borders ($s$ is the stride of the first convolutional layer). Specifically, given a $24 \times 24$ input $\mathbf{Y}_i$ , AR-CNN produces a $(24-s+1)\times (24-s+1)$ output. Hence, the loss (Eqn.~(\ref{eqn:loss})) was computed by comparing against the up-left $(24-s+1)\times (24-s+1)$ pixels of the ground truth sub-image $\mathbf{X}_i$.
In the training phase, we follow~\cite{Jain2009,Dong2014} and use a smaller learning rate ($5 \times 10^{-5}$) in the last layer and a comparably larger one ($5 \times 10^{-4}$) in the remaining layers.

\subsection{Experiments on JPEG-compressed Images}
\label{sec:State-of-the-Arts}


%
We first compare our methods with some state-of-the-art algorithms, including the deblocking oriented method SA-DCT~\cite{Foi2007} and the deep model SRCNN~\cite{Dong2014} and the restoration based RTF~\cite{Jancsary2012}, on restoring JPEG-compressed images.
As in other compression artifacts reduction methods (\eg~RTF~\cite{Jancsary2012}), we apply the standard JPEG compression scheme, and use the JPEG quality settings $q=40,30,20,10$ (from high quality to very low quality) in MATLAB JPEG encoder.
We use the LIVE1 dataset~\cite{Sheikh2005} (29 images) as test set to evaluate both the quantitative and qualitative performance. The LIVE1 dataset contains images with diverse properties. It is widely used in image quality assessment~\cite{Wang2004} as well as in super-resolution~\cite{Yang2014}.
To have a comprehensive qualitative evaluation, we apply the PSNR, structural similarity (SSIM)~\cite{Wang2004}\footnote{We use the unweighted structural similarity defined over fixed $8\times 8$ windows as in~\cite{Wang2008}.}, and PSNR-B~\cite{Yim2011} for quality assessment. We want to emphasize the use of PSNR-B. It is designed specifically to assess blocky and deblocked images.

We use the baseline network settings -- $f_1=9$, $f_{2}=7$, $f_3=1$, $f_4=5$, $n_1=64$, $n_{2}=32$, $n_3=16$ and $n_4=1$, denoted as 64(9)-32(7)-16(1)-1(5) or simply AR-CNN. A specific network is trained for each JPEG quality. Parameters are randomly initialized from a Gaussian distribution with a standard deviation of 0.001.


\subsubsection{Comparison with SA-DCT}

\label{sec:sadct}
\begin{table}\scriptsize
\caption{The average results of PSNR (dB), SSIM, PSNR-B (dB) on the LIVE1 dataset.}\label{tab:sadct}
\vspace{-0.15cm}
\begin{center}
\begin{tabular}{|c|c|c|c|c|}
\hline
 Eval. Mat & Quality & JPEG & SA-DCT & AR-CNN \\

\hline\hline
     & 10 & 27.77 & 28.65 & \textbf{29.13}   \\
PSNR & 20 & 30.07 & 30.81 & \textbf{31.40}   \\
     & 30 & 31.41 & 32.08 & \textbf{32.69}   \\
     & 40 & 32.35 & 32.99 & \textbf{33.63}   \\
\hline\hline
     & 10 & 0.7905 & 0.8093 & \textbf{0.8232}   \\
SSIM & 20 & 0.8683 & 0.8781 & \textbf{0.8886}   \\
     & 30 & 0.9000 & 0.9078 & \textbf{0.9166}   \\
     & 40 & 0.9173 & 0.9240 & \textbf{0.9306}   \\
\hline\hline
     & 10 & 25.33 & 28.01 & \textbf{28.74}   \\
PSNR-B & 20 & 27.57 & 29.82 & \textbf{30.69}  \\
     & 30 & 28.92 & 30.92 & \textbf{32.15}   \\
     & 40 & 29.96 & 31.79 & \textbf{33.12}   \\
\hline
\end{tabular}
\vspace{-0.3cm}
\end{center}
\end{table}

\begin{table}\scriptsize
\caption{The average results of PSNR (dB), SSIM, PSNR-B (dB) on 5 classical test images~\cite{Foi2007}.}\label{tab:sadct2}
\vspace{-0.15cm}
\begin{center}
\begin{tabular}{|c|c|c|c|c|}
\hline
 Eval. Mat & Quality & JPEG & SA-DCT & AR-CNN \\

\hline\hline
     & 10 & 27.82 & 28.88 & \textbf{29.04}   \\
PSNR & 20 & 30.12 & 30.92 & \textbf{31.16}   \\
     & 30 & 31.48 & 32.14 & \textbf{32.52}   \\
     & 40 & 32.43 & 33.00 & \textbf{33.34}   \\
\hline\hline
     & 10 & 0.7800 & 0.8071 & \textbf{0.8111}   \\
SSIM & 20 & 0.8541 & 0.8663 & \textbf{0.8694}   \\
     & 30 & 0.8844 & 0.8914 & \textbf{0.8967}   \\
     & 40 & 0.9011 & 0.9055 & \textbf{0.9101}   \\
\hline\hline
     & 10 & 25.21 & 28.16 & \textbf{28.75}   \\
PSNR-B & 20 & 27.50 & 29.75 & \textbf{30.60}  \\
     & 30 & 28.94 & 30.83 & \textbf{31.99}   \\
     & 40 & 29.92 & 31.59 & \textbf{32.80}   \\
\hline
\end{tabular}
\vspace{-0.3cm}
\end{center}
\end{table}


We first compare AR-CNN with SA-DCT~\cite{Foi2007}, which is widely regarded as the state-of-the-art deblocking oriented method~\cite{Jancsary2012,Li2014}. The quantization results of PSNR, SSIM and PSNR-B are shown in Table~\ref{tab:sadct}. On the whole, our AR-CNN outperforms SA-DCT on all JPEG qualities and evaluation metrics by a large margin. Note that the gains on PSNR-B are much larger than those on PSNR. This indicates that AR-CNN could produce images with less blocking artifacts. We have also conducted evaluation on 5 classical test images used in~\cite{Foi2007}\footnote{The 5 test images in~\cite{Foi2007} are baboon, barbara, boats, lenna and peppers.}, and observed the same trend. The results are shown in Table~\ref{tab:sadct2}.

\begin{figure*}[p]\small
\begin{center}

 \includegraphics[width=\linewidth]{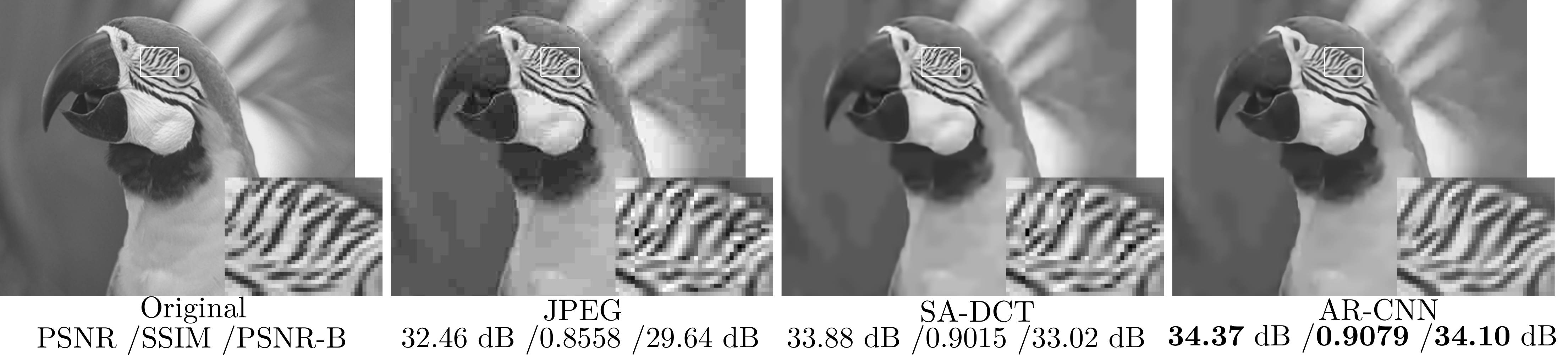}
\caption{Results on image ``parrots'' ($q=10$) show that AR-CNN is better than SA-DCT on removing blocking artifacts.}
\label{fig:sadct1}
\vspace{-0.5cm}
\end{center}
\end{figure*}

To compare the visual quality, we present some restored images with $q=10,20$ in Figure~\ref{fig:sadct1}. 
From the qualitative results, we could see that the result of AR-CNN could produce much sharper edges with much less blocking and ringing artifacts compared with SA-DCT.
The visual quality has been largely improved on all aspects compared with the state-of-the-art method.
Furthermore, AR-CNN is superior to SA-DCT on the implementation speed. For SA-DCT, it needs 3.4 seconds to process a $256\times 256$ image. While AR-CNN only takes 0.5 second. They are all implemented using C++ on a PC with Intel I3 CPU (3.1GHz) with 16GB RAM.

\subsubsection{Comparison with SRCNN}

\label{sec:exp_SRCNN}

\begin{table}\scriptsize
\caption{The average results of PSNR (dB), SSIM, PSNR-B (dB) on the LIVE1 dataset with $q=10$ .}\label{tab:srcnn}
\vspace{-0.15cm}
\begin{center}
\begin{tabular}{|c|c|c|c|c|}
\hline
 Eval. & JPEG & SRCNN  & Deeper & AR-CNN \\
 Mat &       &         & SRCNN &    \\

\hline\hline
PSNR  & 27.77 & 28.91 & 28.92 & \textbf{29.13}   \\
\hline
SSIM & 0.7905 & 0.8175 & 0.8189 & \textbf{0.8232}   \\
\hline
PSNR-B  & 25.33 & 28.52 & 28.46 & \textbf{28.74}   \\
\hline
\end{tabular}
\vspace{-0.3cm}
\end{center}
\end{table}

\begin{figure}[t]\small
\begin{center}
 \includegraphics[width=\linewidth]{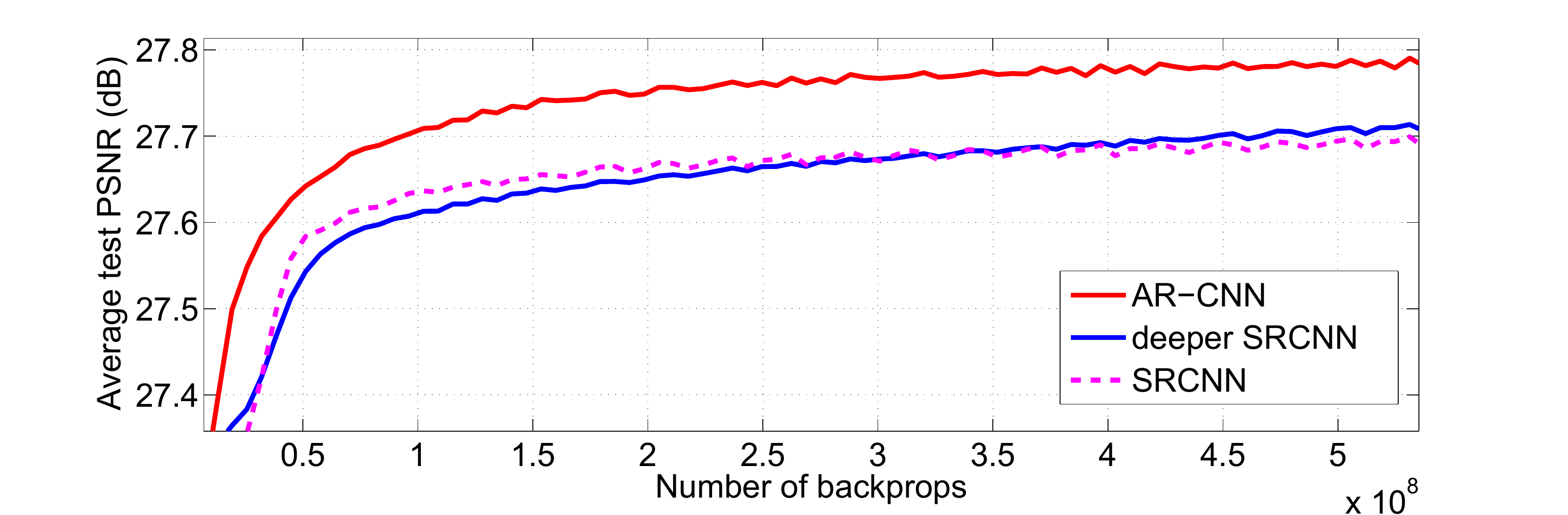}
\caption{Comparisons with SRCNN and Deeper SRCNN.}
\label{fig:srcnn_com}
\vspace{-0.3cm}
\end{center}
\end{figure}

As discussed in Section~\ref{sec:ARCNN}, SRCNN is not suitable for compression artifacts reduction. For comparison, we train two SRCNN networks with different settings.
(i) The original SRCNN (9-1-5) with $f_1=9$, $f_3=5$, $n_1=64$ and $n_2=32$. (ii) Deeper SRCNN (9-1-1-5) with an additional non-linear mapping layer ($f_{3}=1$, $n_{3}=16$).
They all use the BSDS500 dataset for training and validation as in Section~\ref{subsec:settings}. The compression quality is $q=10$.

\begin{figure*}[p]\small
\begin{center}
 \includegraphics[width=\linewidth]{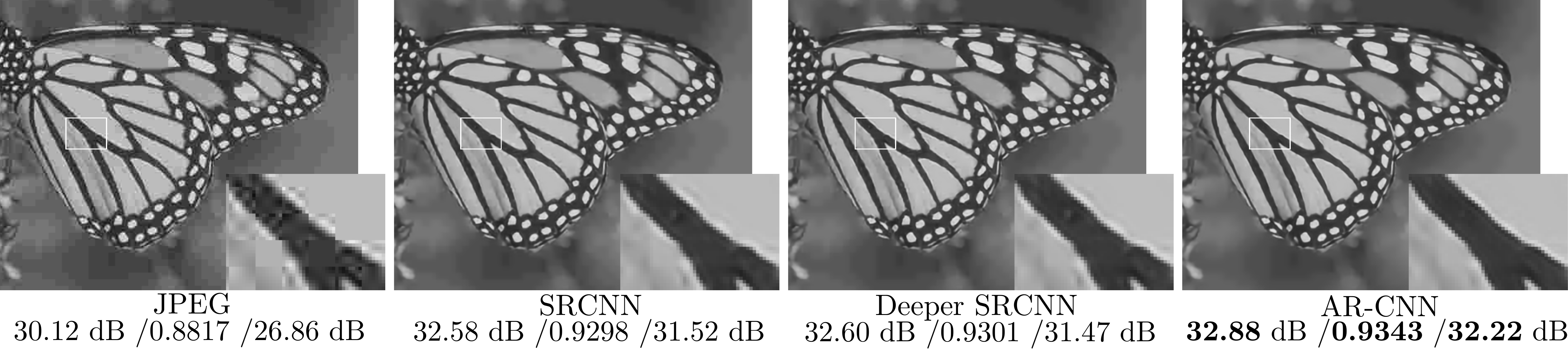}
\caption{Results on image ``monarch'' show that AR-CNN is better than SRCNN on removing ringing effects.}
\label{fig:srcnn}
\vspace{-0.5cm}
\end{center}
\end{figure*}

Quantitative results tested on LIVE1 dataset are shown in Table~\ref{tab:srcnn}. We could see that the two SRCNN networks are inferior on all evaluation metrics. From convergence curves shown in Figure~\ref{fig:srcnn_com}, it is clear that AR-CNN achieves higher PSNR from the beginning of the learning stage.
Furthermore, from their restored images in Figure~\ref{fig:srcnn}, we find out that the two SRCNN networks all produce images with noisy edges and unnatural smooth regions.
These results demonstrate our statements in Section~\ref{sec:ARCNN}. The success of training a deep model needs comprehensive understanding of the problem and careful design of the model structure.

\subsubsection{Comparison with RTF}

\begin{table}\scriptsize
\caption{The average results of PSNR (dB), SSIM, PSNR-B (dB) on the test set BSDS500 dataset.}\label{tab:rtf}
\vspace{-0.15cm}
\begin{center}
\begin{tabular}{|c|c|c|c|c|c|}
\hline
 Eval. & Quality & JPEG & RTF & RTF & AR-CNN \\
Mat &            &      &     & +SA-DCT&    \\
\hline\hline
PSNR & 10 & 26.62 & 27.66 & 27.71 & \textbf{27.79}   \\
     & 20 & 28.80 & 29.84 & 29.87 &\textbf{30.00}   \\
\hline\hline
SSIM & 10 & 0.7904 & 0.8177 & 0.8186 & \textbf{0.8228}   \\
     & 20 & 0.8690 & 0.8864 & 0.8871 & \textbf{0.8899}  \\
\hline\hline
PSNR-B & 10 & 23.54 & 26.93 & 26.99 & \textbf{27.32}   \\
     & 20 & 25.62 & 28.80 & 28.80 & \textbf{29.15}  \\
\hline
\end{tabular}
\vspace{-0.5cm}
\end{center}
\end{table}

RTF~\cite{Jancsary2012} is a recent state-of-the-art restoration oriented method. Without their deblocking code, we can only compare with the released deblocking results. Their model is trained on the training set (200 images) of the BSDS500 dataset, but all images are down-scaled by a factor of 0.5~\cite{Jancsary2012}. To have a fair comparison, we also train new AR-CNN networks on the same half-sized 200 images. Testing is performed on the test set of the BSDS500 dataset (images scaled by a factor of 0.5), which is also consistent with~\cite{Jancsary2012}. We compare with two RTF variants. One is the plain RTF, which uses the filter bank and is optimized for PSNR. The other is the RTF+SA-DCT, which includes the SA-DCT as a base method and is optimized for MAE. The later achieves the highest PSNR value among all RTF variants~\cite{Jancsary2012}.

As shown in Table~\ref{tab:rtf}, we obtain superior performance than the plain RTF, and even better performance than the combination of RTF and SA-DCT, especially under the more representative PSNR-B metric.
Moreover, training on such a small dataset has largely restricted the ability of AR-CNN. The performance of AR-CNN will further improve given more training images.

\begin{figure}[h]\small
\centering
  \includegraphics[width=\linewidth]{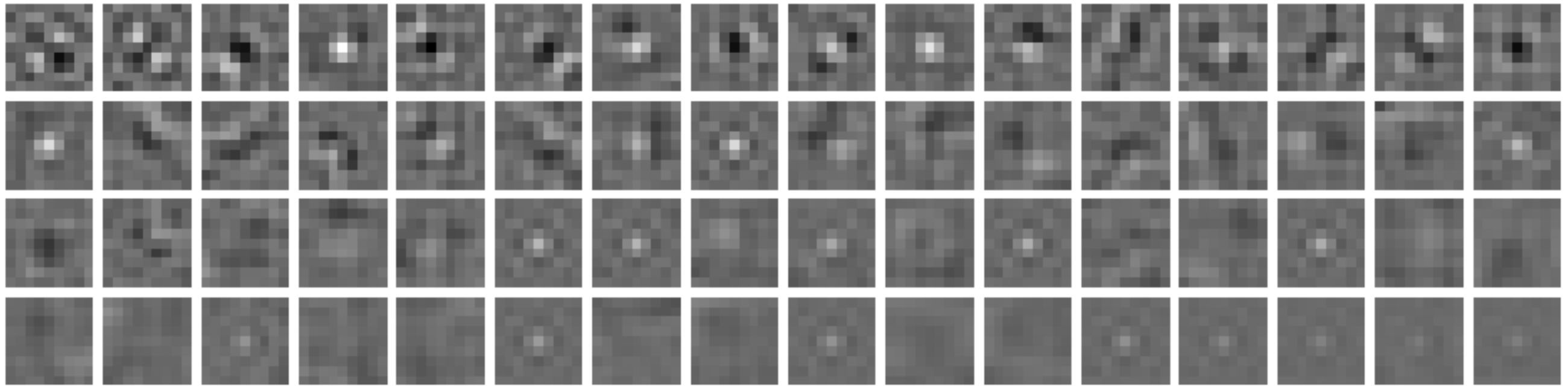}
\caption{First-layer filters of AR-CNN learned for JPEG 2000 at 0.3 BPP.}
\label{fig:feature_j2}
\vskip -0.5cm
\end{figure}

\begin{figure}[h]\small
\centering

  \includegraphics[width=\linewidth]{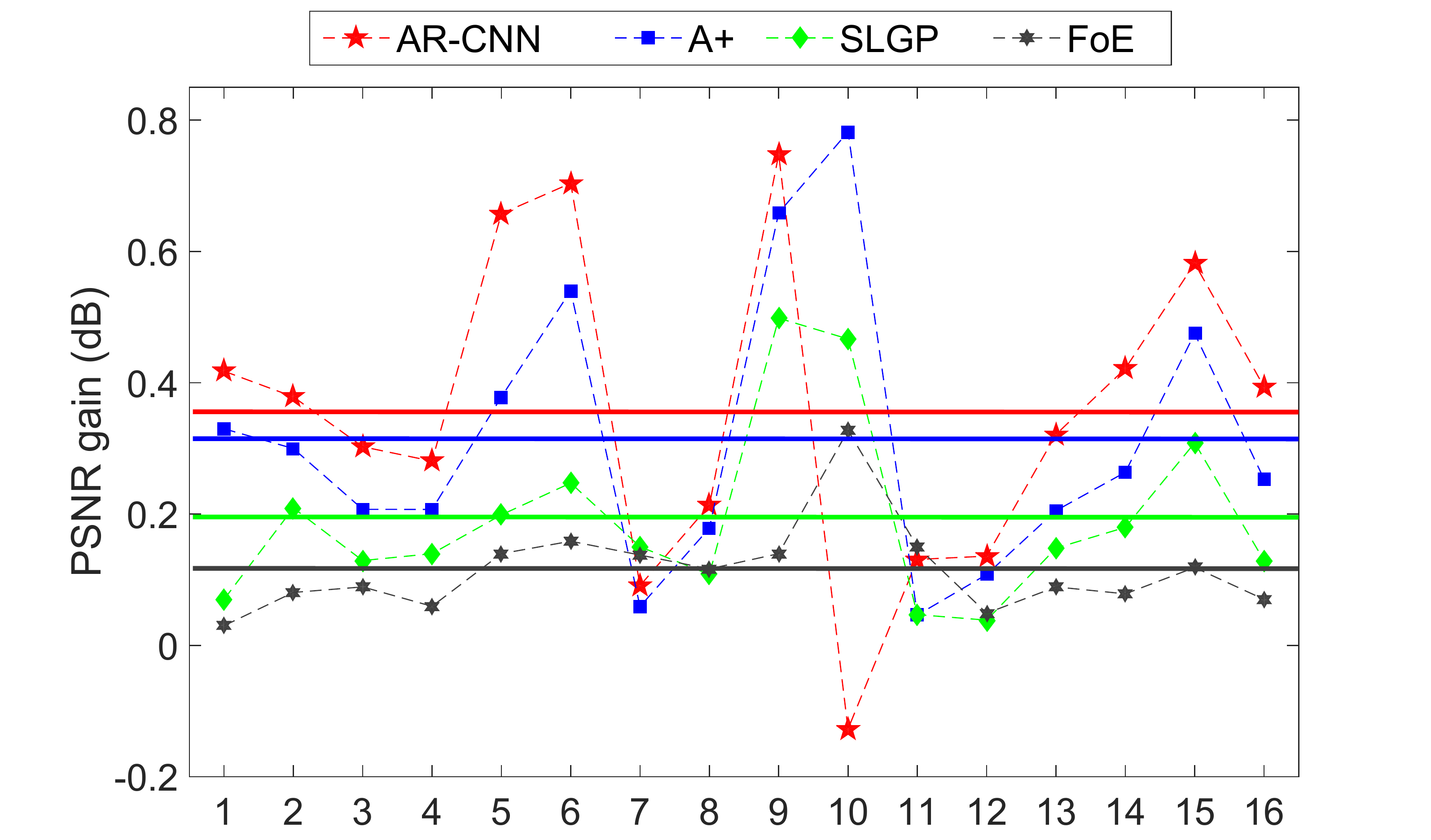}
  \caption{PSNR gain comparison of the proposed AR-CNN
against A+, SLGP , and FoE. The x axis corresponds to the image
index. The average PSNR gains across the
dataset are marked with solid lines.}
\label{compare_j2}
\vskip -0.5cm
\end{figure}

\begin{figure*}[p]\small
\begin{center}
 \includegraphics[width=\linewidth]{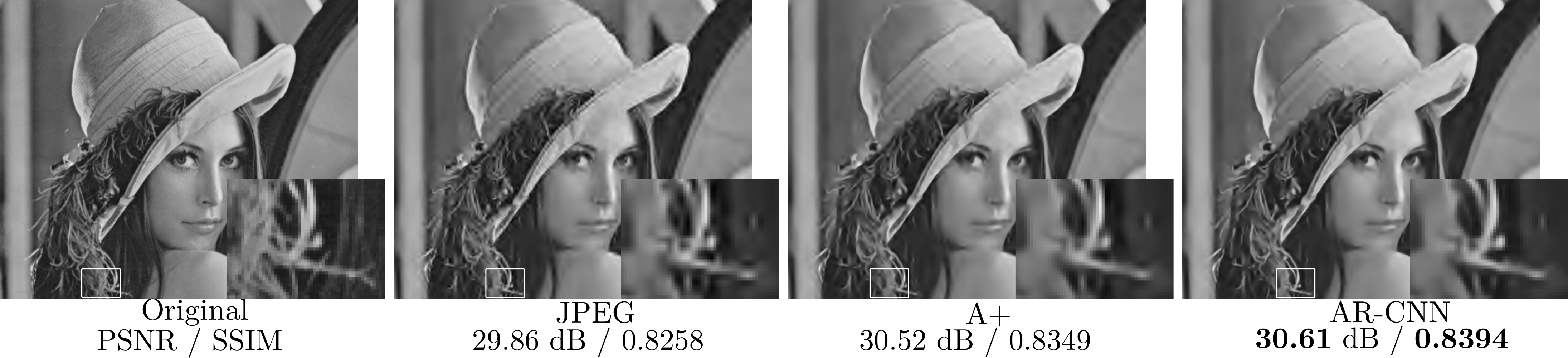}
\caption{Results on image ``lenna'' compressed with JPEG 2000 at 0.1 BPP. }
\label{fig:j2}
\vspace{-0.5cm}
\end{center}
\end{figure*}

\begin{figure*}[p]\small
\begin{center}
 \includegraphics[width=\linewidth]{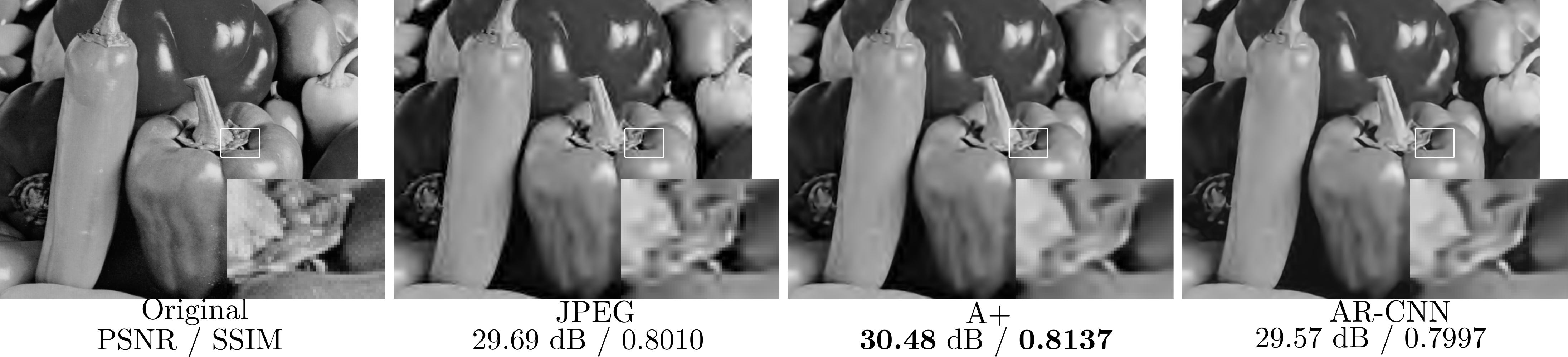}
\caption{Results on image ``pepper'' compressed with JPEG 2000 at 0.1 BPP. }
\label{fig:j2_2}
\vspace{-0.5cm}
\end{center}
\end{figure*}

\begin{figure*}[p]\small
\begin{center}
 \includegraphics[width=\linewidth]{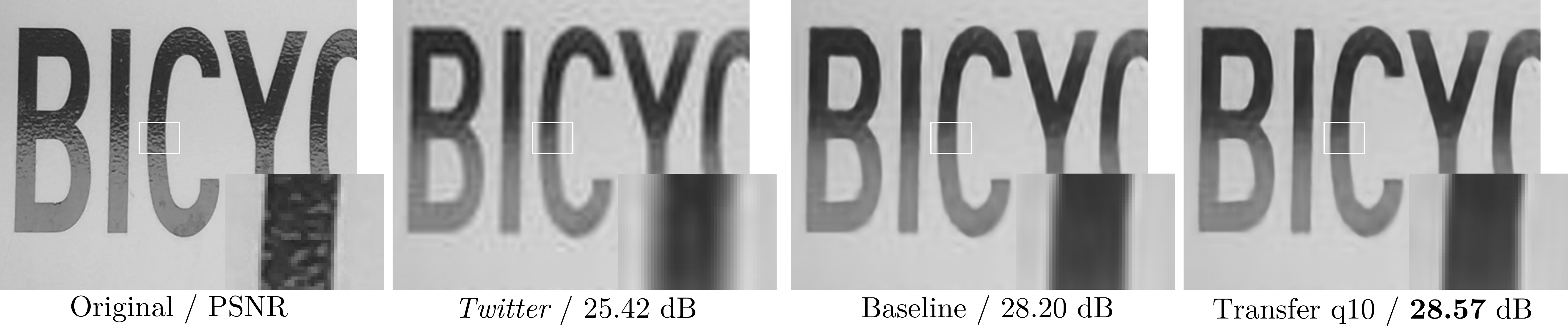}
\caption{Restoration results of AR-CNN on \textit{Twitter}-compressed images.}
\label{fig:twitter}
\vspace{-0.5cm}
\end{center}
\end{figure*}
\subsection{Experiments on JPEG 2000 Images}

As mentioned in the introduction, the proposed AR-CNN is effective in dealing with various compression schemes. In this section, we conduct experiments on the JPEG 2000 standard, and compare with the state-of-the-art method -- the adjusted anchored regression (A+)~\cite{Rothe2015}. To have a fair comparison, we follow A+ on the choice of datasets and software. Specifically, we adopt the the 91-image dataset~\cite{Yang2010a} for training and 16 classical images~\cite{Kwon2015} for testing. The images are compressed using the JPEG 2000 encoder from the Kakadu software package\footnote{\url{http://www.kakadusoftware.com}}.
We also adopt the same training strategy as A+. To test on images degraded at 0.1 bits per pixel (BPP), the training images are compressed at 0.3 BPP instead of 0.1 BPP. As indicated in~\cite{Rothe2015}, the regressors can more easily pick up the artifact patterns at a lower compression rate, leading to better performance. We use the same AR-CNN network structure (64(9)-32(7)-16(1)-1(5)) as in the JPEG experiments. Figure~\ref{fig:feature_j2} shows the patterns of the learned first-layer filters, which differ a lot from that for JPEG images (see Figure~\ref{fig:features}).

Apart from A+, we compare our results against another two methods -- SLGP~\cite{Kwon2015} and FoE~\cite{Roth2009}. The PSNR gains of the 16 test images are shown in Figure~\ref{compare_j2}. It is observed that our method outperforms others on most test images. For the average performance, we achieve a PSNR gain of 0.353 dB, better than A+ with 0.312 dB, SLGP with 0.192 dB and FoE with 0.115 dB. Note that the improvement is already significant in such a difficult scenario -- JPEG 2000 at 0.1 BPP~\cite{Rothe2015}. Figure~\ref{fig:j2} shows some qualitative results, where our method achieves better PSNR and SSIM than A+. However, we also notice that AR-CNN is inferior to other methods on the tenth image in Figure~\ref{compare_j2}. The restoring results of this image are shown in Figure~\ref{fig:j2_2}. It is observed that the result of AR-CNN is still visually pleasant, and the lower PSNR is mainly due to the chromatic aberration in smooth regions. The above experiments demonstrate the generalization ability of AR-CNN on handling different compression standards.

During training, we also find that AR-CNN is hard to converge using random initialization mentioned in Section~\ref{sec:State-of-the-Arts}. We solve the problem by adopting the transfer learning strategy. To be specific, we can transfer the first-layer filters of a well-trained three-layer network to the four-layer AR-CNN, or we can reuse the features of AR-CNN trained on the JPEG images. They refer to different `easy-hard transfer'' strategies -- \textit{transfer shallow to deeper model} and \textit{transfer standard to real use case}, which will be detailed in the following section.


\subsection{Experiments on Easy-Hard Transfer}
\label{sec:transfer}

We show the experimental results of different ``easy-hard transfer'' settings on JPEG-compressed images. The details of settings are shown in Table~\ref{tab:transfer}. Take the base network as an example, the ``base-q10'' is a four-layer AR-CNN 64(9)-32(7)-16(1)-1(5) trained on the BSDS500~\cite{Arbelaez2011} dataset (400 images) under the compression quality $q=10$. Parameters are initialized by randomly drawing from a Gaussian distribution with zero mean and standard deviation 0.001. Figures~\ref{fig:transfer1}~-~\ref{fig:transfer3} show the convergence curves on the validation set.
%

\begin{table}\scriptsize
\caption{Experimental settings of ``easy-hard transfer''. The ``9-7-1-5'' and ``9-7-3-1-5'' are short for 64(9)-32(7)-16(1)-1(5) and 64(9)-32(7)-16(3)-16(1)-1(5), respectively.}\label{tab:transfer}
\vspace{-0.15cm}
\begin{center}
\begin{tabular}{|c|c|c|c|c|c|}
\hline
transfer & short & network   & training  & initialization \\
strategy & form  & structure & dataset   & strategy       \\
\hline\hline
base     & base-q10 & 9-7-1-5 & BSDS-q10  & Gaussian (0, 0.001)  \\
network  & base-q20 & 9-7-1-5 & BSDS-q20  & Gaussian (0, 0.001)  \\
\hline\hline
shallow  & base-q10 & 9-7-1-5 & BSDS-q10  & Gaussian (0, 0.001)  \\
to       & transfer deeper & 9-7-3-1-5 & BSDS-q10  & 1,2 layers of base-q10\\
deep   & He~\cite{He2015} & 9-7-3-1-5 & BSDS-q10  & He~\etal~\cite{He2015}\\
\hline\hline
high  & base-q10 & 9-7-1-5 &BSDS-q10  & Gaussian (0, 0.001)  \\
to       & transfer 1 layer & 9-7-1-5 & BSDS-q10  & 1 layer of base-q20\\
low   & transfer 2 layers & 9-7-1-5 &BSDS-q10  & 1,2 layer of base-q20 \\
\hline\hline
standard  & base-Twitter & 9-7-1-5 & \textit{Twitter}  & Gaussian (0, 0.001)  \\
to       & transfer q10 & 9-7-1-5 & \textit{Twitter}  & 1 layer of base-q10\\
real   & transfer q20 & 9-7-1-5 & \textit{Twitter} & 1 layer of base-q20 \\
\hline
\end{tabular}
\end{center}
\end{table}

\subsubsection{Transfer shallow to deeper model}
\label{sec:transfer1}

\begin{figure}[t]\small
\begin{center}
 \includegraphics[width=\linewidth]{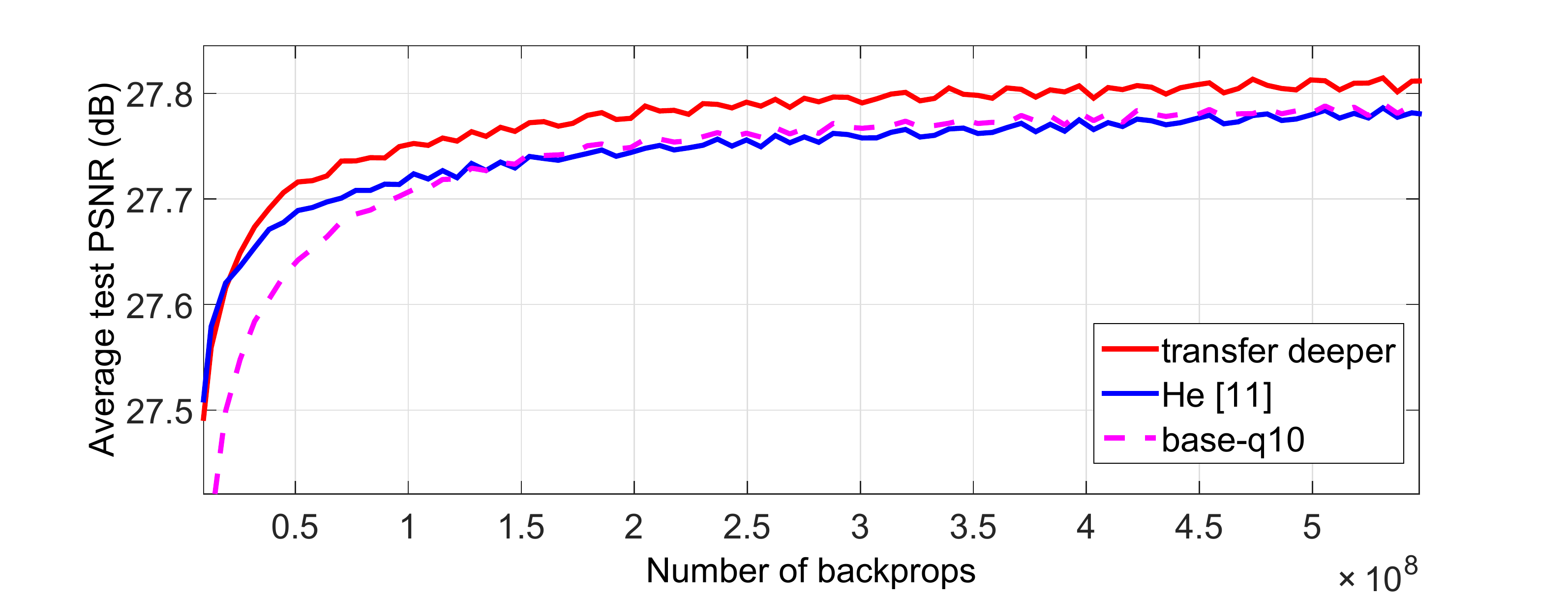}
\vskip -0.15cm
\caption{Transfer shallow to deeper model.}
\label{fig:transfer1}
\vspace{-0.5cm}
\end{center}
\end{figure}

In Table~\ref{tab:transfer}, we denote a deeper (five-layer) AR-CNN 64(9)-32(7)-16(3)-16(1)-1(5) as ``9-7-3-1-5''.
Results in Figure~\ref{fig:transfer1} show that the transferred features from a four-layer network enable us to train a five-layer network successfully. Note that directly training a five-layer network using conventional initialization ways is unreliable. Specifically, we have exhaustively tried different groups of learning rates, but still could not observe convergence. Furthermore, the ``transfer deeper'' converges faster and achieves better performance than using He~\etal's method~\cite{He2015}, which is also very effective in training a deep model. We have also conducted comparative experiments with the structure 64(9)-32(7)-16(1)-16(1)-1(5) and 64(9)-32(1)-32(7)-16(1)-1(5), and observed the same trend.

\subsubsection{Transfer high to low quality}
\label{sec:transfer2}
\begin{figure}[t]\small
\begin{center}
 \includegraphics[width=\linewidth]{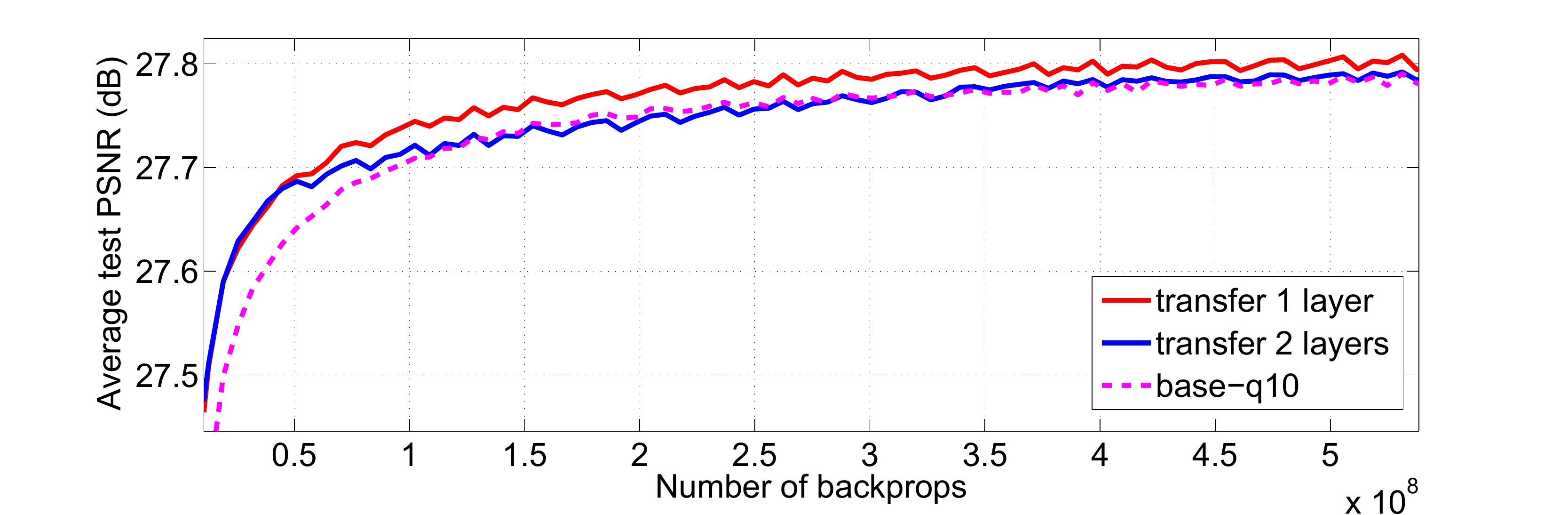}
\vskip -0.15cm
\caption{Transfer high to low quality.}
\label{fig:transfer2}
\vspace{-0.3cm}
\end{center}
\end{figure}

\begin{figure}[t]\small
\begin{center}
 \includegraphics[width=\linewidth]{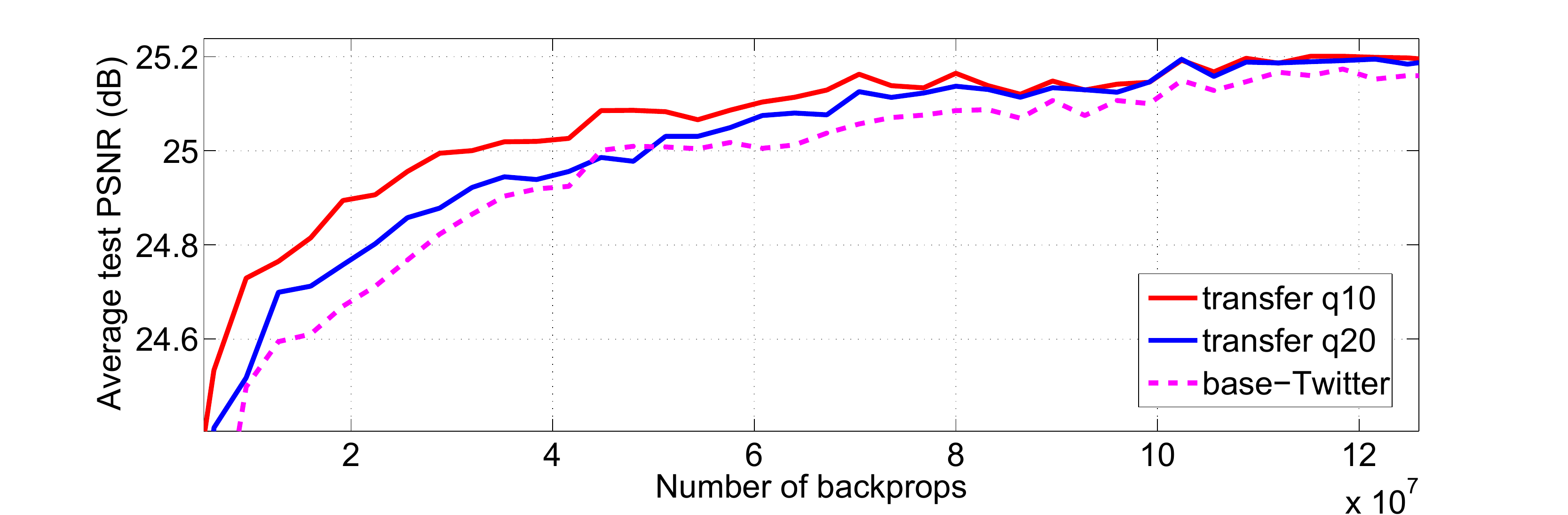}
\vskip -0.15cm
\caption{Transfer standard to real use case.}
\label{fig:transfer3}
\vspace{-0.5cm}
\end{center}
\end{figure}

Results are shown in Figure~\ref{fig:transfer2}. Obviously, the two networks with transferred features converge faster than that training from scratch. For example, to reach an average PSNR of 27.77dB, the ``transfer 1 layer'' takes only $1.54\times 10^8$ backprops, which are roughly a half of that for ``base-q10''. Moreover, the ``transfer 1 layer'' also outperforms the `transfer 2 layers'' by a slight margin throughout the training phase. One reason for this is that only initializing the first layer provides the network with more flexibility in adapting to a new dataset.
This also indicates that a good starting point could help train a better network with higher convergence speed.

\subsubsection{Transfer standard to real use case -- \textit{Twitter}}
\label{sec:transfer3}

Online Social Media like \textit{Twitter} are popular platforms for message posting. However, \textit{Twitter} will compress the uploaded images on the server-side. For instance, a typical 8 mega-pixel (MP) image ($3264 \times 2448$) will result in a compressed and re-scaled version with a fixed resolution of $600 \times 450$. Such re-scaling and compression will introduce very complex artifacts, making restoration difficult for existing deblocking algorithms (\eg~SA-DCT). However, AR-CNN can fit to the new data easily. Further, we want to show that features learned under standard compression schemes could also facilitate training on a completely different dataset.
We use 40 photos of resolution $3264 \times 2448$ taken by mobile phones (totally 335,209 training subimages) and their \textit{Twitter}-compressed version\footnote{We have shared this dataset on our project page \url{http://mmlab.ie.cuhk.edu.hk/projects/ARCNN.html}.} to train three networks with initialization settings listed in Table~\ref{tab:transfer}.

From Figure~\ref{fig:transfer3}, we observe that the ``transfer $q10$'' and ``transfer $q20$'' networks converge much faster than the ``base-Twitter'' trained from scratch. Specifically, the ``transfer $q10$'' takes $6\times 10^7$ backprops to achieve 25.1dB, while the ``base-Twitter'' uses $10\times 10^7$ backprops. Despite of fast convergence, transferred features also lead to higher PSNR values compared with ``base-Twitter''. This observation suggests that features learned under standard compression schemes are also transferrable to tackle real use case problems.
Some restoration results are shown in Figure~\ref{fig:twitter}. We could see that both networks achieve satisfactory quality improvements over the compressed version.

\subsection{Experiments on Acceleration Strategies}

In this section, we conduct a set of controlled experiments to demonstrate the effectiveness of the proposed acceleration strategies. Following the descriptions in Section~\ref{subsec:Acceleration}, we progressively modify the baseline AR-CNN by layer decomposition, adopting large-stride layers and expanding the mapping layer. The networks are trained on JPEG images under the quality $q=10$.  We further test the performance of Fast AR-CNN on different compression qualities ($q=10,20,30,40$). As all the modified networks are deeper than the baseline model, we adopt the proposed transfer learning strategy (transfer shallow to deeper model) for fast and stable training. The base network is also ``base-q10'' as in Section~\ref{sec:transfer1}. All the quantitative results are listed in Table~\ref{tab:settings}.

\subsubsection{Layer decomposition}
\label{sec:settings1}

The layer decomposition strategy replaces the ``feature enhancement'' layer with a ``shrinking'' layer and an ``enhancement'' layer, and we reach to a modified network 64(9)-32(1)-32(7)-16(1)-1(5). The experimental results are shown in Table~\ref{tab:settings}, from which we can see that the ``replace deeper'' achieves almost the same performance as the ``base-q10'' in all the metrics. This indicates that the layer decomposition is an effective strategy to reduce the network parameters with almost no performance loss.

\subsubsection{Stride size}

\begin{figure}[t]\small
\begin{center}
 \includegraphics[width=\linewidth]{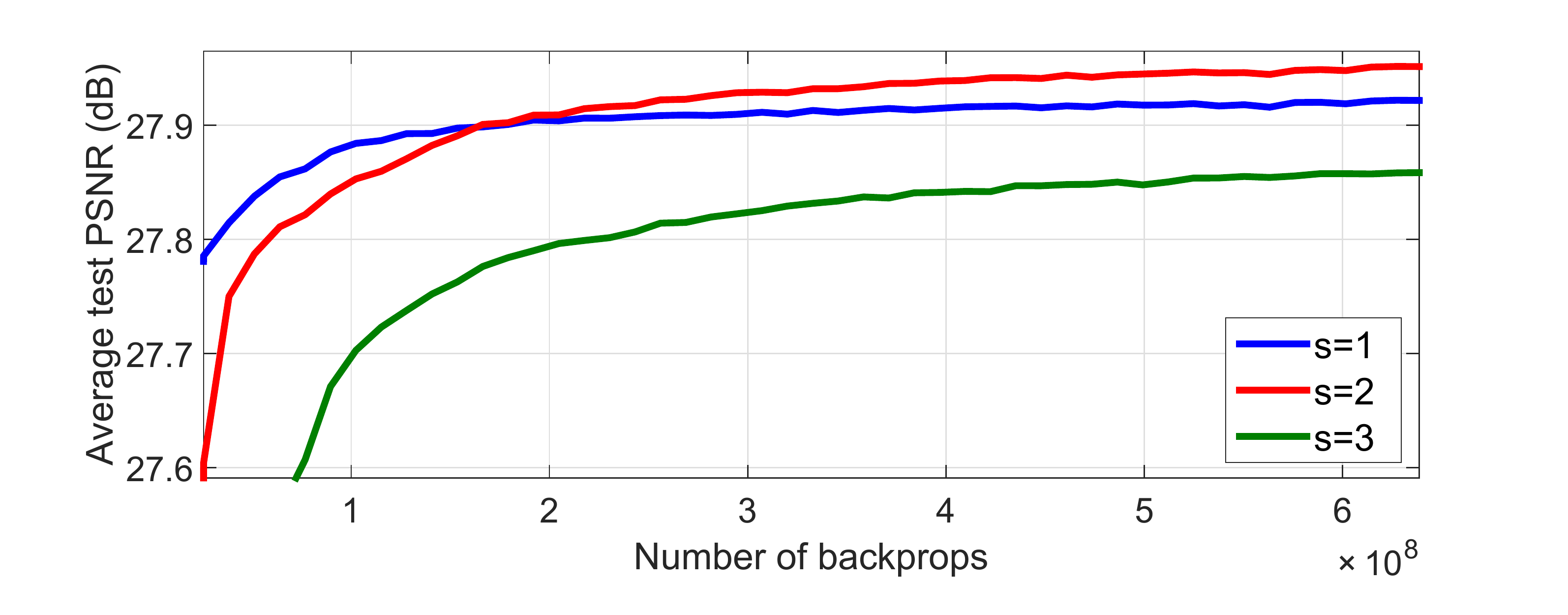}
\vskip -0.15cm
\caption{The performance of using different stride sizes.}
\label{fig:stride}
\vspace{-0.5cm}
\end{center}
\end{figure}

Then we introduce the large-stride convolutional and deconvolutional layers, and change the stride size. Generally, a larger stride will lead to much narrower feature maps and faster inference, but at the risk of worse reconstruction quality. In order to find a good trade-off setting, we conduct experiments with different stride sizes as shown in the part ``stride'' of Table~\ref{tab:settings}. The network settings for $s=1$, $s=2$ and $s=3$ are 64(9)-32(1)-32(7)-16(1)-1(5), 64(9)-32(1)-32(7)-16(1)-1[9]-s2 and 64(9)-32(1)-32(7)-16(1)-1[9]-s3, respectively. From the results in Table~\ref{tab:settings}, we can see that there are only small differences between ``$s=1$'' and ``$s=2$'' in all metrics. But when we further enlarge the stride size, the performance declines dramatically, ~\eg~the PSNR value drops more than 0.2 dB from ``$s=2$'' to ``$s=3$''.
Convergence curves in Figure~\ref{fig:stride} also exhibit a similar trend, where ``$s=3$'' achieves inferior performance to ``$s=1$'' and ``$s=2$'' on the validation set\footnote{As the validation set (BSD500 validation set) is different from the test set (LIVE1 dataset), the results in Table~\ref{tab:settings} and Figure~\ref{fig:stride} are different.}.
With little performance loss yet 7.5 times faster, using stride $s=2$ definitely balances the performance and time complexity. Thus we adopt stride $s=2$ in the following experiments.

\begin{table}\scriptsize
\caption{The experimental results of different settings.}\label{tab:settings}
\vspace{-0.15cm}
\begin{center}
\begin{tabular}{|c|c|c|c|c|}
\hline
\multicolumn{2}{|c|}{Eval. Mat}  & PSNR(dB)      & SSIM    & PSNR-B(dB) \\
\hline\hline
layer        & base-q10               & \textbf{29.13} & 0.8232 & \textbf{28.74}  \\
replacement  & replace deeper        & 29.13 & \textbf{0.8234} & 28.72  \\
\hline\hline
         & $s=1$  & \textbf{29.13}  & \textbf{0.8234} & \textbf{28.72}  \\
stride   & $s=2$  & 29.07  & 0.8232 & 28.66  \\
         & $s=3$  & 28.78           & 0.8178 & 28.44  \\
\hline\hline
         & $n_4=16$  & 29.07  & 0.8232 & 28.66  \\
mapping  & $n_4=48$  & 29.04  & 0.8238 & 28.58  \\
filters  & $n_4=64$  & \textbf{29.10}  & \textbf{0.8246} & 28.65  \\
         & $n_4=80$  & 29.10  & 0.8244 & \textbf{28.69}  \\
\hline\hline

 & fast-q10   & 29.10  & \textbf{0.8246} & 28.65  \\
 & base-q10 & \textbf{29.13}  & 0.8232 & \textbf{28.74}  \\
\cline{2-5}
     & fast-q20   & 31.29  & 0.8873 & 30.54  \\
JPEG & base-q20 & \textbf{31.40}  & \textbf{0.8886} & \textbf{30.69}  \\
\cline{2-5}
quality & fast-q30   & 32.41  & 0.9124 & 31.43  \\
        & base-q30 & \textbf{32.69}  & \textbf{0.9166} & \textbf{32.15}  \\
\cline{2-5}
& fast-q40   & 33.43  & 0.9306 & 32.51  \\
& base-q40 & \textbf{33.63}  & \textbf{0.9306} & \textbf{33.12}  \\
\hline
\end{tabular}
\end{center}
\end{table}

\subsubsection{Mapping filters}
\begin{figure}[t]\small
\begin{center}
 \includegraphics[width=\linewidth]{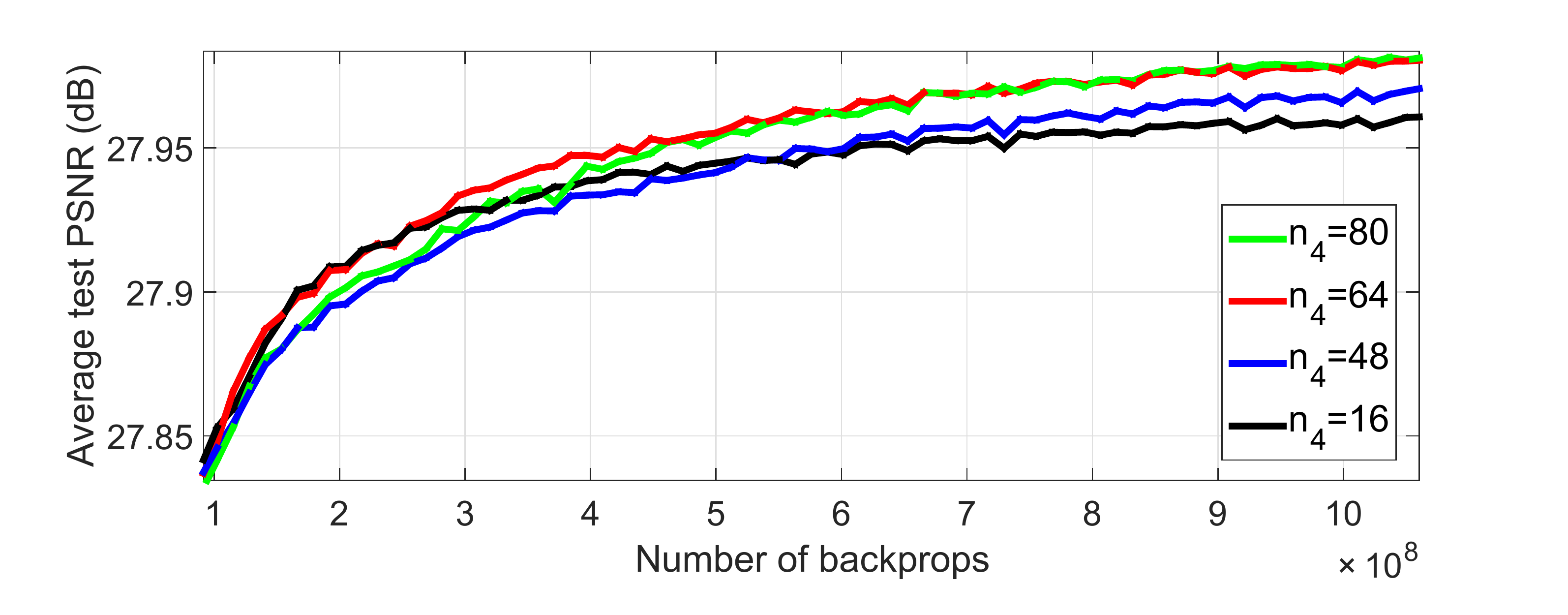}
\vskip -0.25cm
\caption{The performance of using different mapping filters.}
\label{fig:mapping}
\vspace{-0.5cm}
\end{center}
\end{figure}

As mentioned in Section~\ref{subsec:Acceleration}, we can increase the number of mapping filters to compensate the performance loss. In the part ``mapping filters'' of Table~\ref{tab:settings}, we compare a set of experiments that only differ in mapping filters. To be specific, the network setting is 64(9)-32(1)-32(7)-$n_4$(1)-1[9]-s2 with $n_4=16,48,64,80$. The convergence curves shown in Figure~\ref{fig:mapping}\footnote{As the validation set (BSD500 validation set) is different from the test set (LIVE1 dataset), the results in Table~\ref{tab:settings} and Figure~\ref{fig:mapping} are different.}. can better reflect their differences. Obviously, using more filters will achieve better performance, but the improvement is marginal beyond $n_4=64$. Thus we adopt $n_4=64$, which is also consistent with our comment in Section~\ref{subsec:Acceleration}. Finally, we find the optimal network setting -- 64(9)-32(1)-32(7)-64(1)-1[9]-s2, namely Fast AR-CNN, which achieves similar performance as the baseline model 64(9)-32(7)-16(1)-1(5) but is 7.5 times faster.

\subsubsection{JPEG quality}
In the above experiments, we mainly focus on a very low quality $q=10$. Here we want to examine the capacity of the new network on different compression qualities. In the part ``JPEG quality'' of Table~\ref{tab:settings}, we compare the Fast AR-CNN with the baseline AR-CNN on quality $q=10,20,30,40$. For example, ``fast-q10'' and ``base-q10'' represent 64(9)-32(1)-32(7)-64(1)-1[9]-s2 and 64(9)-32(7)-16(1)-1(5) on quality $q=10$, respectively. From the quantitative results, we observe that the Fast AR-CNN is comparable with AR-CNN on low qualities such as $q=10$ and $q=20$, but it is inferior to AR-CNN on high qualities such as $q=30$ and $q=40$. This phenomenon is reasonable. As the low quality images contain much less information, extracting features in a sparse way (using a large stride) does little harm to the restoration quality. On the contrary, for high quality images, adjacent image patches may differ a lot. So when we adopt a large stride, we will lose the information that is useful for restoration. Nevertheless, the proposed Fast AR-CNN still outperforms the state-of-the-art methods (as presented in Section \ref{sec:State-of-the-Arts}) on different compression qualities.

\section{Conclusion}

Applying deep model on low-level vision problems requires deep understanding of the problem itself. In this paper, we carefully study the compression process and propose a four-layer convolutional network, AR-CNN, which is extremely effective in dealing with various compression artifacts. Then we propose two acceleration strategies to reduce its time complexity while maintaining good performance. We further systematically investigate three \textit{easy-to-hard} transfer settings that could facilitate training a deeper or better network, and verify the effectiveness of transfer learning in low-level vision problems.

\ifCLASSOPTIONcaptionsoff
  \newpage
\fi

\bibliographystyle{ieee}
\bibliography{long.bib,DCNN.bib}

\end{document}